\documentclass[11pt]{article}
\usepackage[hyphens]{url}   %
\usepackage[table]{xcolor}		
\usepackage[preprint]{acl}

\usepackage{CJKutf8}

\usepackage[nott]{newtxtext}	%
\usepackage{newtxmath}
\usepackage{latexsym}

\usepackage{dblfloatfix}
\usepackage{capt-of}  %

\usepackage[T1]{fontenc}
\usepackage[utf8]{inputenc}

\usepackage{microtype}

\usepackage{enumitem} %
\newlist{paradesc}{description}{1}
\setlist[paradesc]{style=unboxed,leftmargin=0pt,parsep=\parskip,listparindent=\parindent,labelsep=1em}

\usepackage{graphicx}
\usepackage{booktabs}   %
\usepackage{multirow} %
\usepackage{siunitx}
\NewCommandCopy{\oldCellcolor}{\cellcolor}
\RenewDocumentCommand{\cellcolor}{O{named} m}{\oldCellcolor[#1]{#2}\ignorespaces}

\usepackage{calc}       %
\usepackage{ifthen}     %
\usepackage{relsize} %
\usepackage{needspace}  %
\def\textsubscript#1{\ensuremath{_{\mbox{\textscale{0.7}{#1}}}}}

\newcommand{\pstars}[2]{%
    \ifthenelse{\equal{#2}{{---}}}{#2}{$#2$}%
    \parbox[t]{\widthof{$^{\ast\ast\ast}$}}{$^{%
        \ifthenelse{\equal{#1}{-}}{}{%
            \ifthenelse{\equal{#1}{}}{}{%
                \ifthenelse{\equal{#1}{*}}{\ast}{%
                    \ifthenelse{\equal{#1}{**}}{\ast\ast}{%
                        \ast\ast\ast%
                    }%
                }%
            }%
        }%
        }$\hfill%
    }%
}

\newcommand{\TableGroup}[2]{\multirow[c]{#1}{*}{\makebox[6pt][l]{\rotatebox[origin=c]{90}{#2}}}}
\NewDocumentCommand{\TableItem}{O{0pt} m}{%
    \parbox[t]{\linewidth+#1}{%
        --
        \parbox[t]{\linewidth-\widthof{-- }+#1}{%
            \raggedright{#2}%
        }%
    }\hspace{-#1}%
}

\defcitealias{ninjal_2016_construction}{NINJAL}
\defcitealias{ninjal_2018_wordlist}{NINJAL}
\defcitealias{bncconsortium_2007_british}{BNC Consortium}
\defcitealias{realacademiaespanola_2004_corpus}{Real Academia Española}
\defcitealias{vanparidon_thompson_2021_subs2vec}{Van Paridon and Thompson} %
\newcommand{\citealpaliasyear}[1]{\citetalias{#1}, \citeyear{#1}}
\newcommand{\citepaliasyear}[1]{(\citealpaliasyear{#1})}
\newcommand{\citetaliasyear}[1]{\citetalias{#1} (\citeyear{#1})}
\usepackage{xspace}

\newcommand{\Tdefault}{TUBE\-LEX\textsubscript{default}\xspace}

\newcommand{\Tbase}{TUBE\-LEX\textsubscript{base}\xspace}
\newcommand{\Tlemma}{TUBE\-LEX\textsubscript{lemma}\xspace}

\title{Beyond Film Subtitles: Is YouTube the Best Approximation of Spoken Vocabulary?}

\author{%
    \begin{tabular}{c<{\hspace{-1.5em}}c<{\hspace{-1.5em}}c}
        \parbox{\widthof{Eunike Andriani Kardinata}}{\centering Adam Nohejl} & 
        Frederikus Hudi &
        Eunike Andriani Kardinata\\
        Shintaro Ozaki &
        Maria Angelica Riera Machin &
        Hongyu Sun\\
    &\mbox{\hspace{-5em}Justin Vasselli \hspace{6em}
        Taro Watanabe\hspace{-5em}}&
    \end{tabular}\\
        Nara Institute of Science and Technology\\
    $\{\texttt{nohejl.adam.mt3},\;\texttt{frederikus.hudi.fe7},\;\texttt{eunike.kardinata.ef9}\}\texttt{@is.naist.jp}$\\
    $\{\texttt{ozaki.shintaro.ou6},\;\texttt{riera\_machin.maria.rn9},\;\texttt{sun.hongyu.sg6}\}\texttt{@naist.ac.jp}$\\
    $\{\texttt{vasselli.justin\_ray.vk4},\;\texttt{taro}\}\texttt{@is.naist.jp}$
    }

\begin{document}
\maketitle

\begin{abstract}
Word frequency is a key variable in psycholinguistics, useful for modeling human familiarity with words even in the era of large language models (LLMs). Frequency in film subtitles has proved to be a particularly good approximation of everyday language exposure. For many languages, however, film subtitles are not easily available, or are overwhelmingly translated from English. We demonstrate that frequencies extracted from carefully processed YouTube subtitles provide an approximation comparable to, and often better than, the best currently available resources. Moreover, they are available for languages for which a high-quality subtitle or speech corpus does not exist. We use YouTube subtitles to construct frequency norms for five diverse languages, Chinese, English, Indonesian, Japanese, and Spanish, and evaluate their correlation with lexical decision time, word familiarity, and lexical complexity. In addition to being strongly correlated with two psycholinguistic variables, a simple linear regression on the new frequencies achieves a new high score on a lexical complexity prediction task in English and Japanese, surpassing both models trained on film subtitle frequencies and the LLM GPT-4. Our code, the frequency lists, fastText word embeddings, and statistical language models are freely available online.\footnote{
    \url{https://github.com/naist-nlp/tubelex} %
}
\end{abstract}

\section{Introduction}

Word frequency is crucial for psycholinguistic research, as well as for assistive or educational applications involving production or comprehension of words.  

Psycholinguistic analyses of the relative strength of variables affecting lexical processing, such as word frequency and age of acquisition (e.g.\ \citealp{garlock_etal_2001_age}), hinge on accurate data for these variables. As word frequency and age of acquisition are correlated with each other, having less representative frequency data can easily change the result of such an analysis.

Traditionally, written corpora have been used for estimates of word frequency, with \citet{kucera_francis_1967_computational} frequency norms long dominating psycholinguistic research of English. While the size of written language corpora has grown over time, speech corpora are still costly to develop and comparably limited in extent. When user-generated text became available on a large scale, it was possible to approximate everyday language exposure by collecting English text from Usenet newsgroups \citep{burgess_livesay_1998_effect}, and later French film and TV subtitles \citep{new_etal_2007_use}. Subtitle-based norms for US English, SUBTLEX-US, \citep{brysbaert_new_2009_moving} were found more predictive of lexical decision times (LDT) than frequencies based on traditional written corpora or the Usenet-based corpus.

These pioneering studies on subtitle corpora spurred the creation of film and TV subtitle-based frequency norms, dubbed SUBTLEX, for other languages such as Spanish \citep{cuetos_etal_2011_subtlex},
or British English \citep{vanheuven_etal_2014_subtlex}. %
SUBTLEX frequencies for two Asian languages, Chinese \citep{cai_brysbaert_2010_subtlex} and Vietnamese \citep{pham_etal_2019_constructing} were compiled as well. Most of the research, however, has remained focused on languages spoken in WEIRD\footnote{Western, Educated, Industrial, Rich, and Democratic (WEIRD), an acronym coined by \citet{henrich_2020_weirdest}.} countries.

Subtitle frequencies are currently being used in a variety of practical tasks which need to model familiarity with words, such as lexical simplification or readability assessment. %
Despite their practical utility, film subtitle corpora are far from perfect approximations of spoken language. A large part of the non-English SUBTLEX corpora comes from translations of English-language movies. For instance, SUBTLEX-ESP \citep{cuetos_etal_2011_subtlex} consists of less than 3\% original Spanish subtitles, while more than 92\% are translations from English. 
In Vietnamese, \citet{pham_etal_2019_constructing} did not find subtitles more predictive of LDT than a written corpus, citing translation artifacts and cultural differences from predominantly American material as the likely causes. %
Moreover, the content presented in film dialogue is a very specific subset of spoken language.
The speech is almost exclusively scripted and skewed to particular topics and vocabulary \citep{paetzold-specia-2016-collecting}.

In this work, we build a corpus of untranslated YouTube video subtitles and evaluate the correlation of its frequencies with LDT, word familiarity, and lexical complexity, comparing them with frequencies based on available subtitle and speech corpora. We purposely target two languages spoken in WEIRD countries, English and Spanish, with a wealth of previous research to compare with, as well as three languages with diverse characteristics and amounts of resources available, Chinese, Japanese, and Indonesian.

As full corpus data cannot be published due to copyright, we release two basic language models based on the TUBELEX corpus for each language in addition to the frequency lists:
a statistical language model \citep{heafield-etal-2013-scalable}, which provides smoothed frequencies of word 1-grams to 5-grams, and
fastText word embeddings \citep{bojanowski-etal-2017-enriching} to enable modeling of semantic similarity or analogy, as well as representation of words in downstream application. FastText extends the Word2vec model \citep{mikolov_etal_2013_efficient}.
Preprocessing details and hyperparameters are provided in \autoref{sec:model-settings}, model sizes in \autoref{sec:model-sizes}, and evaluation of the embeddings in \autoref{sec:eval-embed}.

\section{Related Work}
\label{sec:related}

\subsection{Subtitle Corpora}

\citet{new_etal_2007_use} collected French film subtitles from the web %
to create a subtitle corpus. A similar procedure was then used for SUBTLEX-US \cite{brysbaert_new_2009_moving}, and other SUBTLEX corpora, in some cases adding duplicate removal, e.g.\ for SUBTLEX-ESP \citep{cuetos_etal_2011_subtlex}, or various forms of cleaning. While most of the film subtitle corpora are collected from the web (often the OpenSubtitles website\footnote{\url{http://www.opensubtitles.org/}}), the British SUBTLEX-UK \citep{vanheuven_etal_2014_subtlex} acquired television subtitles from the BBC broadcasts.

\citet{francom-etal-2014-activ} used film metadata to build a relatively small corpus of untranslated Spanish subtitles, ACTIV-ES, %
and released lists of its $n$-grams. \citet{paetzold-specia-2016-collecting} restricted movies and series to particular genres to build the SubIMDB corpus.

All of these corpora aim to approximate spoken language and most of them were evaluated against psycholinguistic data. Other subtitle corpora were built for different purposes:

OpenSubtitles2016 \citep{lison-tiedemann-2016-opensubtitles2016} and its updated version OpenSubtitles2018 \citep{lison-etal-2018-opensubtitles2018} are large-scale collections of parallel film and TV subtitles downloaded from the OpenSubtitles website. In addition to parallel text aligned via subtitle timing,  word frequencies for individual languages were released as well. %
For some languages, such as Indonesian, the OpenSubtitles corpus is the only subtitle corpus available.

\citet{takamichi_etal_2021_jtubespeech} downloaded audio and subtitles from YouTube to create JTubeSpeech, a Japanese corpus for speech recognition and speaker verification. The corpus or derived data was not published, and the corpus was evaluated only on these two tasks.

\subsection{Evaluation Methods and Applications}
\label{sec:eval-app}

\citet{new_etal_2007_use} evaluated a French subtitle corpus using correlation with LDT to demonstrate that it reflects language exposure better than written corpora. The same approach was subsequently adopted by others for different languages.
\citet{paetzold-specia-2016-collecting} additionally evaluated the English SubIMDB on four other psycholinguistic ratings including word familiarity. 

\citetaliasyear{vanparidon_thompson_2021_subs2vec} used the data from several OpenSubtitles2018 languages to train word embeddings and evaluated them on word analogy and psycholinguistic ratings. %
The study excluded Chinese, Japanese, and other languages that do not separate words with spaces.

\citet{shardlow-2013-comparison} demonstrated that frequency in SUBTLEX-US outperforms frequency in written corpora in ranking lexical simplifications for native speakers. Subtitle frequencies have been widely applied to lexical simplification in various languages for native and non-native speakers, where suitable SUBTLEX corpora were available (e.g.\ \citealp{stajner_etal_2022_lexical}). %
Meanwhile, lexical complexity modeling for other languages, such as Japanese \citep{nishihara-kajiwara-2020-word} or Indonesian \citep{wibowo_etal_2019_lexical}, has had to rely on web-scraped corpora instead.

Subtitle frequencies have also been used in a number of other tasks broadly connected to text comprehension, assistive technologies, and language learning, e.g.\ text readability assessment in English \citep{chen-meurers-2016-characterizing} and Italian \citep{okinina-etal-2020-ctap}, modeling of the orthographic neighborhood effect in English and Dutch \citep{tulkens-etal-2020-orthographic}, a cross-linguistic study of the mental lexicon in English, German, and Chinese \citep{tjuka-2020-general}, construction of a vocabulary list for Finnish language learners \citep{robertson-etal-2022-tallvocabl2fi}, or evaluating and improving performance of LLMs in colloquial English \citep{sun-etal-2024-toward}.

\section{Corpus Construction} %
\label{sec:construction}

We build the corpus using several stages of processing. \autoref{tab:stats} shows statistic of the process.

\subsection{Subtitle Scraping}

As there is no public index of YouTube videos, we use YouTube's search function to search for all Wikipedia article titles in a given language to discover videos, following \citet{takamichi_etal_2021_jtubespeech}.

To avoid translated or machine-generated subtitles, we restrict videos to those with both audio and manual subtitles explicitly labeled as the target language. For Chinese videos, we did not find enough videos with labeled audio language, so we also accept videos with unlabeled audio. The resulting numbers of videos are listed as Found in \autoref{tab:stats}. We sample 120,000 videos for each language, for which we download subtitles.

\subsection{Cleaning and Duplicate Removal}
\label{sec:cleaning}

We identify the language of each subtitle line using the compressed fastText language identification model\footnote{\url{https://fasttext.cc/docs/en/language-identification.html}} \citep{joulin_etal_2016_fasttext,joulin_etal_2016_bag}.
We discard files containing less than 95\% of the target language. From the remaining files,
we remove both lines that do not contain any valid characters for the target language (e.g.\ Latin alphabet for English), and lines that are identified as a different language. Lastly, we discard any files less than three lines long. The resulting numbers of files  (i.e.\ files not discarded during cleaning out of the sample of size 120,000) are listed as Cleaned in \autoref{tab:stats}.

We consider files duplicate if the cosine similarity between their 1-gram TF-IDF vectors is 0.95 or higher. We remove duplicate files heuristically to achieve a state without any duplicate pair. The final numbers of unique files and tokens in them are listed as Unique and Tokens in \autoref{tab:stats}.

\begin{table}[t]
    \footnotesize
    \setlength\tabcolsep{3.3pt}
    \centering

\begin{tabular}{lrrrr}
\toprule
\multirow{2}{*}{Language} &
\multicolumn{3}{c}{Videos} &
\multirow{2}{*}{Tokens\textsuperscript{\ddag}}\\
\cmidrule{2-4}
 & Found & Cleaned\textsuperscript{\dag} & Unique\textsuperscript{\dag} &  \\
\midrule
Chinese & 5,848,257 & 10,172 & 10,146 & 17,865,686 \\
English & 4,748,327 & 105,976 & 105,752 & 170,750,870 \\
Indonesian & 5,265,240 & 34,818 & 34,684 & 34,903,381 \\
Japanese & 4,970,247 & 101,664 & 100,754 & 163,439,781 \\
Spanish & 3,840,068 & 107,166 & 106,676 & 169,188,689 \\
\bottomrule
\end{tabular}
\caption{
    Corpus construction statistics.
    \textsuperscript{\dag}Out of 120,000 downloaded subtitle files.
    \textsuperscript{\ddag}In default tokenization.
}
\label{tab:stats}
\end{table}

\subsection{Subtitle Processing}

We parse the WebVTT\footnote{\url{https://www.w3.org/TR/webvtt1/}} subtitle files, and remove formatting and repetition caused by subtitle scrolling. We preserve words censored by YouTube (replaced with ``[~\_{\kern 1pt}\_~]'')\footnote{\url{https://support.google.com/youtube/answer/6373554?hl=en}} and audio descriptions in brackets (e.g.\ English ``[ominous music]'', Japanese ``\textsmaller{{\kern -0.4em}\begin{CJK}{UTF8}{min}【エンジン音】\end{CJK}{\kern -0.4em}}'') as special tokens.

\subsection{Masking Personal Information}

We also use special tokens to replace sequences of digits (after tokenization) and anonymize email addresses, web addresses including those without an explicit protocol (e.g.\ \texttt{x.com/username}), and apparent social network handles starting with \texttt{@}. Our approach to anonymization is informed by the analysis by \citet{subramani-etal-2023-detecting} and extends the approach of \citet{soldaini_etal_2024_dolma} by masking web addresses and social network handles.

\subsection{Tokenization and Frequency Lists}
\label{sec:tok-freq-lists}

We provide frequency lists in multiple variants:

\begin{paradesc}
\item[default] English, Indonesian, and Spanish segmented using Stanza \citep{qi-etal-2020-stanza} \texttt{tokenize,\allowbreak{}mwt} pipeline%
; Japanese segmented using MeCab \citep{kudo-etal-2004-applying} with UniDic 2.1.2 (\citealp{den_etal_2007_development}, distributed as unidic-lite\footnote{\url{https://pypi.org/project/unidic-lite/}}); Chinese segmented using the jieba\footnote{\url{https://github.com/fxsjy/jieba}} segmenter 0.42.1.

\item[base] Base form of Japanese tokens, preserving original spelling, obtained from MeCab/UniDic (\textsmaller{\begin{CJK}{UTF8}{min}書字形基本形\end{CJK}}).

\item[lemma] English, Indonesian and Spanish lemmatized using Stanza \texttt{tokenize,mwt,lemma} pipeline; Japanese lemmatized using MeCab/Unidic (\textsmaller{\begin{CJK}{UTF8}{min}語彙素\end{CJK}}), i.e. words in the orthographically normalized base form. %

\item[regex] English, Indonesian, and Spanish orthographic words, matching a Python regular expression for sequences of characters belonging to the \verb|\w| (word) class, but not to the \verb|\d| (digit) class. %

\end{paradesc}

\label{sec:human-eval}
\begin{table*}[t]
    \footnotesize
    \setlength\tabcolsep{6pt}
    \centering

\begin{tabular}{clrrrrr}
\toprule
Status & Evaluation Result & Chinese & English & Indonesian & Japanese & Spanish \\
\midrule
OK & Subtitles match speech in the target language & 65\% & 91\% & 84\% & 84\% & 89\% \\
\hline
\multirow[c]{6}{*}{\makebox[6pt][l]{\rotatebox[origin=c]{90}{Problematic}}} & Subtitles match song in the target language & ---\phantom{\%} & 2\% & 1\% & 1\% & 2\% \\
& Subtitles provide audio description (e.g. \textit{phone rings}) & ---\phantom{\%} & 1\% & 1\% & 0\% & 0\% \\
&Audio is neither speech or song & ---\phantom{\%} & 2\% & 0\% & 0\% & 2\% \\
& Audio is synthesized speech & 34\% & 3\% & 13\% & 14\% & 5\% \\
& Audio language differs from the target language & 1\% & 1\% & 1\% & 1\% & 2\% \\
& Subtitle language differs from the target language & ---\phantom{\%} & ---\phantom{\%} & ---\phantom{\%} & ---\phantom{\%} & ---\phantom{\%} \\
\bottomrule
\end{tabular}

    \caption{Human evaluation of a sample size 300 for each language, consisting of 100 videos with 3 cues per video. Each of the 300 subtitle cues per language was assigned to exactly one of the evaluation results.%
    }
    \label{tab:human-eval}
\end{table*}

All tokens are lower-cased and normalized to Unicode NFKC \citep{whistler_2023_uax}. For each word, the frequency lists provide: \textbf{\texttt{count}} -- number of occurrences, \textbf{\texttt{videos}} -- number of videos containing the word, \textbf{\texttt{channels}} -- number of channels the word occurrs in, \textbf{\texttt{count:}}$C$ -- number of occurrences of the word in the YouTube video category~$C$. %

\section{Human Evaluation}

We performed human evaluation to verify how representative the corpus is of the target languages, and spoken language in particular. We took a sample of size 300 for each language, consisting of 100 videos with 3 random subtitle time stamps for each. %
The videos were selected using stratified sampling by category and duration\footnote{
We divided the videos in three similarly large duration classes: [0, 3~min), [3~min, 10~min), and [10~min, $\infty$).
} for each language.
Each sampled timestamp was examined and labeled by a CEFR C2-level non-native speaker for English and by native speakers for the other languages. We examined each example by playing the video and comparing the subtitle cue (the subtitle text displayed at the given timestamp) with the corresponding audio, extending the examined video segment as deemed necessary by the evaluator to categorize the example.

The evaluation results are shown in \autoref{tab:human-eval}. To be evaluated as ``OK'', the subtitles must match human speech in audio and both must be in the target language. Otherwise, the instance is assigned to one of the ``Problematic'' evaluation results. While in principle problems not listed in the table may occur (e.g.\ the language matches, but the content of audio and subtitles differs) or the listed problems could co-occur, it did not happen in our sample.

Most importantly, among the 1,500 subtitle cues, we have not found a single one whose language would differ from the target language.
We have, however, observed different dialects or varieties of each language, as well as apparent non-native speech, sometimes co-occurring in the same video.
In the case of Chinese, 2 of the 100 videos were in Cantonese, with traditional Chinese subtitles, while the majority was in Mandarin Chinese. To better understand the composition of the Chinese subtitles, we also analyzed the script used in the whole corpus, and found that 66\% videos use simplified Chinese, 33\% videos use traditional Chinese, and 1\% mix both.\footnote{
    We used the Hanzi Identifier package (\url{https://github.com/tsroten/hanzidentifier}). %
    and considered subtitles mixed if they contained the non-majority script variant on at least 2 lines and at least 5\% of lines.
}

Most proportions of potentially problematic phenomena were relatively low (up to 2\%), with the exception of synthesized speech, which ranged from 3\% for English to 34\% for Chinese.
Synthesized speech with subtitles, or subtitles provided for scenes without speech, could effectively be written language, rather than spoken.
We further discuss the implications in \autoref{sec:spoken-vs-written}.

\section{Extrinsic Evaluation}
\label{sec:ext-eval}

We evaluate multiple corpora on three tasks, LDT, word familiarity, and lexical complexity, comparing them with TUBELEX in default, base, lemma, and regex variants, described in \autoref{sec:tok-freq-lists}.

Note that the datasets available for different languages generally have different characteristics (such as part of speech or word frequency distribution), so while our experiments allow comparison of different corpora for a particular task and language, they do not allow comparison across languages. For instance, we should not surmise that TUBELEX provides better data for Spanish than for English (or that lexical decision time is generally more strongly correlated with frequency in Spanish than in English) only because TUBELEX achieves (or because all corpora achieve) a higher correlation for the task in Spanish than English.%

For each evaluated corpus, we report correlation measured by Pearson's correlation coefficient (PCC), and the statistical significance of its difference from the correlation with \Tdefault on three levels: $^{\ast\ast\ast}$ ($p < 0.001$), $^{\ast\ast}$ ($p < 0.01$), and $^{\ast}$ ($p < 0.05$). We compute the $p$-values using \citeauthor{steiger_1980_tests}'s (\citeyear{steiger_1980_tests}) test for dependent correlations and consider $p \ge 0.05$ not statistically significant.

To demonstrate the practical usefulness of the TUBELEX frequencies, we also predict lexical complexity based on them, and compare our results with the top submissions of the BEA 2024 Multilingual Lexical Simplification Pipeline Shared Task \citep{shardlow-etal-2024-bea}. %

\subsection{Evaluated Corpora and Resources}
\label{sec:eval-corpora}

We evaluate traditional speech corpora, subtitle corpora, and three additional resources:

\begin{paradesc}

\item[Speech corpora:]
\textbf{BNC-Spoken}, the spoken subset of the British National Corpus \citepaliasyear{bncconsortium_2007_british}; \textbf{CREA-Spo\-ken}, the spoken subset of Corpus de Referencia del Español Actual \citepaliasyear{realacademiaespanola_2004_corpus}; \textbf{CSJ}, the Corpus of Spontaneous Japanese  \citepaliasyear{ninjal_2016_construction}; \textbf{HKUST/MTS} \citep{liu_etal_2006_hkust}, a Mandarin telephone speech corpus. We could not find a large enough Indonesian speech corpus.

\item[Subtitle corpora:]
\textbf{ACTIV-ES};
\textbf{EsPal} \citep{duchon_etal_2013_espal};
\textbf{LaboroTV1+2}, the combination of the two releases of LaboroTVSpeech \citep{ando_fujihara_2021_construction};
\textbf{OpenSubtitles}, the 2018 version;
\textbf{SubIMDB};
\textbf{SUBTLEX} (US, CH, ESP);
\textbf{SUBTLEX-UK}.

\item[Other resources:] %
\textbf{GINI}, a Twitter-based metric of words' dispersion in frequency of use by different people (\citealp{murayama_etal_2018_word}; also see \autoref{sec:gini});
\textbf{Wikipedia};
\textbf{wordfreq}, a Python library \citep{robyn_speer_2022_wordfreq} pooling frequency from multiple corpora. Wordfreq combines Wikipedia, Twitter and a subtitle corpus for each of the evaluated languages, as well as 4 more sources for English, Chinese, and Spanish, and 2 more for Japanese. The subtitle data used by wordfreq is OpenSubtitles2018, SUBTLEX-US and SUBTLEX-UK for English, SUBTLEX-CH for Chinese. %
\end{paradesc}

For each corpus, we provide technical details in \autoref{sec:corpora-details}, and token and type counts in \autoref{sec:stats-corpora}. %

\subsection{Computing Frequency}
\label{sec:frequency}

To deal with words missing in a corpus, we use the formula with Laplace smoothing recommended by \citet{brysbaert_diependaele_2013_dealing} to compute frequency of a token $w$:
\begin{equation}
\label{eq:frequency}
f(w) = \frac{\textrm{count}(w) + 1}{\textrm{\#tokens} + \textrm{\#types}},
\end{equation}
where $\textrm{count}(w)$ is the number of occurrences of the word $w$, \#tokens is the total number of tokens in the corpus, and \#types is the number of types in the corpus. As a result, even words missing in the corpus are assigned a non-zero frequency. %
In all experiments we use the logarithm of frequency. \autoref{sec:corpora-details} provides details about specific corpora.

\subsection{Lexical Decision Time}
\label{sec:ldt}

Lexical decision is one of the basic psycholinguistic tasks, where subjects decide whether a sequence of characters is a valid word or not. The reaction time for each word is its lexical decision time (LDT).

We measure correlation (PCC) with mean LDT from three studies: the English Lexicon Project
\citep{balota_etal_2007_english}, restricted to lower-case words following the approach of \citet{brysbaert_new_2009_moving}; the MELD-SCH database \citep{tsang_etal_2018_meld} of simplified Chinese words%
; and SPALEX \citep{aguasvivas_etal_2018_spalex} for Spanish.
For English and Chinese, we use the published mean LDT. SPALEX only provides raw participant data, which we process by removing times out of the range [200~ms, 2000~ms], as outlined by \citet{aguasvivas_etal_2018_spalex}, and computing the means. 

\begin{table}[t]
    \setlength\tabcolsep{3.5pt}
    \footnotesize
    \centering
    \begin{tabular}{llccc}
\toprule
{} & {Corpus} & {Chinese} & {English} & {Spanish} \\
\midrule
\multirow[c]{3}{*}{\makebox[6pt][l]{\rotatebox[origin=c]{90}{speech}}} & BNC-Spoken & \pstars{-}{{---}} & {\cellcolor[HTML]{56A0CE}} \color[HTML]{F1F1F1} \pstars{***}{-0.548} & \pstars{-}{{---}} \\
 & CREA-Spoken & \pstars{-}{{---}} & \pstars{-}{{---}} & {\cellcolor[HTML]{CDE0F1}} \color[HTML]{000000} \pstars{***}{-0.645} \\
 & HKUST/MTS & {\cellcolor[HTML]{C4DAEE}} \color[HTML]{000000} \pstars{***}{-0.465} & \pstars{-}{{---}} & \pstars{-}{{---}} \\
\hline
\multirow[c]{6}{*}{\makebox[6pt][l]{\rotatebox[origin=c]{90}{\vphantom{l}\textsmaller{film/TV subtitles}}}} & ACTIV-ES & \pstars{-}{{---}} & \pstars{-}{{---}} & {\cellcolor[HTML]{F7FBFF}} \color[HTML]{000000} \pstars{***}{-0.600} \\
 & EsPal & \pstars{-}{{---}} & \pstars{-}{{---}} & {\cellcolor[HTML]{083573}} \color[HTML]{F1F1F1} \pstars{***}{-0.807} \\
 & OpenSubtitles & {\cellcolor[HTML]{084F99}} \color[HTML]{F1F1F1} \pstars{}{-0.568} & {\cellcolor[HTML]{08306B}} \color[HTML]{F1F1F1} \pstars{***}{\mathbf{-0.647}} & {\cellcolor[HTML]{08306B}} \color[HTML]{F1F1F1} \pstars{}{-0.811} \\
 & SubIMDB & \pstars{-}{{---}} & {\cellcolor[HTML]{08306B}} \color[HTML]{F1F1F1} \pstars{***}{-0.646} & \pstars{-}{{---}} \\
 & SUBTLEX & {\cellcolor[HTML]{08306B}} \color[HTML]{F1F1F1} \pstars{**}{\mathbf{-0.587}} & {\cellcolor[HTML]{084082}} \color[HTML]{F1F1F1} \pstars{**}{-0.633} & {\cellcolor[HTML]{1C6BB0}} \color[HTML]{F1F1F1} \pstars{***}{-0.763} \\
 & SUBTLEX-UK & \pstars{-}{{---}} & {\cellcolor[HTML]{084990}} \color[HTML]{F1F1F1} \pstars{}{-0.625} & \pstars{-}{{---}} \\
\hline
\multirow[c]{3}{*}{\makebox[6pt][l]{\rotatebox[origin=c]{90}{other}}} & GINI & \pstars{-}{{---}} & {\cellcolor[HTML]{F7FBFF}} \color[HTML]{000000} \pstars{***}{-0.420} & \pstars{-}{{---}} \\
 & Wikipedia & {\cellcolor[HTML]{F6FAFF}} \color[HTML]{000000} \pstars{***}{-0.424} & {\cellcolor[HTML]{63A8D3}} \color[HTML]{F1F1F1} \pstars{***}{-0.540} & {\cellcolor[HTML]{6DAFD7}} \color[HTML]{F1F1F1} \pstars{***}{-0.705} \\
 & wordfreq & {\cellcolor[HTML]{F7FBFF}} \color[HTML]{000000} \pstars{***}{-0.423} & {\cellcolor[HTML]{084184}} \color[HTML]{F1F1F1} \pstars{**}{-0.632} & {\cellcolor[HTML]{083C7D}} \color[HTML]{F1F1F1} \pstars{***}{-0.801} \\
\hline
\multirow[c]{3}{*}{\makebox[6pt][l]{\rotatebox[origin=c]{90}{our\vphantom{l}}}} & TUBELEX\textsubscript{default} & {\cellcolor[HTML]{084488}} \color[HTML]{F1F1F1} \pstars{-}{-0.575} & {\cellcolor[HTML]{08468B}} \color[HTML]{F1F1F1} \pstars{-}{-0.627} & {\cellcolor[HTML]{08306B}} \color[HTML]{F1F1F1} \pstars{-}{-0.811} \\
 & TUBELEX\textsubscript{regex} & \pstars{-}{{---}} & {\cellcolor[HTML]{08468B}} \color[HTML]{F1F1F1} \pstars{}{-0.627} & {\cellcolor[HTML]{08306B}} \color[HTML]{F1F1F1} \pstars{}{\mathbf{-0.811}} \\
 & TUBELEX\textsubscript{lemma} & \pstars{-}{{---}} & {\cellcolor[HTML]{084A91}} \color[HTML]{F1F1F1} \pstars{*}{-0.624} & {\cellcolor[HTML]{083471}} \color[HTML]{F1F1F1} \pstars{***}{-0.808} \\
\bottomrule
\end{tabular}

    \caption{LDT correlation. Strongest (lowest) correlations for each language are in bold.}
    \label{tab:ldt}
\end{table}

\begin{figure}[t]
    \centering
    \includegraphics[width=1.0\linewidth]{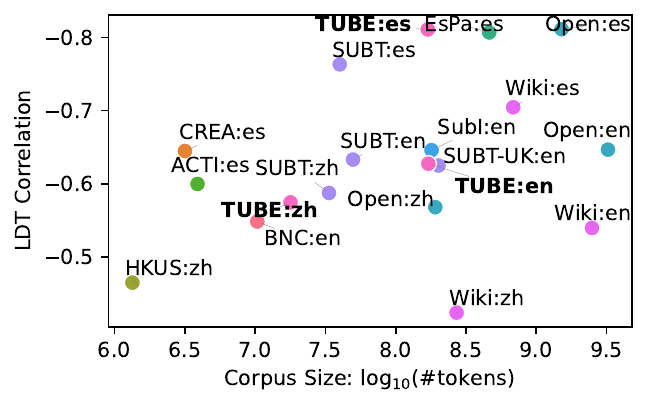}
    \caption{LDT correlation and corpus size. Labeled ``\textit{corpus abbr.}:\textit{lang.\ code}'', ``TUBE'' is \Tdefault.}
    \label{fig:ldt-size}
\end{figure}

\begin{table*}[t]
    \footnotesize
    \centering
    \begin{tabular}{llccccc}
\toprule
{} & {Corpus} & {Chinese} & {English} & {Indonesian} & {Japanese} & {Spanish} \\
\midrule
\multirow[c]{4}{*}{\makebox[6pt][l]{\rotatebox[origin=c]{90}{speech}}} & BNC-Spoken & \pstars{-}{{---}} & {\cellcolor[HTML]{3888C1}} \color[HTML]{F1F1F1} \pstars{***}{0.741} & \pstars{-}{{---}} & \pstars{-}{{---}} & \pstars{-}{{---}} \\
 & CREA-Spoken & \pstars{-}{{---}} & \pstars{-}{{---}} & \pstars{-}{{---}} & \pstars{-}{{---}} & {\cellcolor[HTML]{08468B}} \color[HTML]{F1F1F1} \pstars{}{0.535} \\
 & CSJ & \pstars{-}{{---}} & \pstars{-}{{---}} & \pstars{-}{{---}} & {\cellcolor[HTML]{3E8EC4}} \color[HTML]{F1F1F1} \pstars{***}{0.523} & \pstars{-}{{---}} \\
 & HKUST/MTS & {\cellcolor[HTML]{3B8BC2}} \color[HTML]{F1F1F1} \pstars{***}{0.414} & \pstars{-}{{---}} & \pstars{-}{{---}} & \pstars{-}{{---}} & \pstars{-}{{---}} \\
\hline
\multirow[c]{7}{*}{\makebox[6pt][l]{\rotatebox[origin=c]{90}{film/TV subtitles}}} & ACTIV-ES & \pstars{-}{{---}} & \pstars{-}{{---}} & \pstars{-}{{---}} & \pstars{-}{{---}} & {\cellcolor[HTML]{084F99}} \color[HTML]{F1F1F1} \pstars{}{0.526} \\
 & EsPal & \pstars{-}{{---}} & \pstars{-}{{---}} & \pstars{-}{{---}} & \pstars{-}{{---}} & {\cellcolor[HTML]{82BBDB}} \color[HTML]{000000} \pstars{***}{0.428} \\
 & LaboroTV1+2 & \pstars{-}{{---}} & \pstars{-}{{---}} & \pstars{-}{{---}} & {\cellcolor[HTML]{1D6CB1}} \color[HTML]{F1F1F1} \pstars{***}{0.565} & \pstars{-}{{---}} \\
 & OpenSubtitles & {\cellcolor[HTML]{1E6DB2}} \color[HTML]{F1F1F1} \pstars{***}{0.444} & {\cellcolor[HTML]{083979}} \color[HTML]{F1F1F1} \pstars{}{0.776} & {\cellcolor[HTML]{2A7AB9}} \color[HTML]{F1F1F1} \pstars{***}{0.582} & {\cellcolor[HTML]{F7FBFF}} \color[HTML]{000000} \pstars{***}{0.314} & {\cellcolor[HTML]{08306B}} \color[HTML]{F1F1F1} \pstars{}{\mathbf{0.553}} \\
 & SubIMDB & \pstars{-}{{---}} & {\cellcolor[HTML]{08306B}} \color[HTML]{F1F1F1} \pstars{}{\mathbf{0.781}} & \pstars{-}{{---}} & \pstars{-}{{---}} & \pstars{-}{{---}} \\
 & SUBTLEX & {\cellcolor[HTML]{08316D}} \color[HTML]{F1F1F1} \pstars{}{0.505} & {\cellcolor[HTML]{084184}} \color[HTML]{F1F1F1} \pstars{}{0.773} & \pstars{-}{{---}} & \pstars{-}{{---}} & {\cellcolor[HTML]{084285}} \color[HTML]{F1F1F1} \pstars{}{0.538} \\
 & SUBTLEX-UK & \pstars{-}{{---}} & {\cellcolor[HTML]{083471}} \color[HTML]{F1F1F1} \pstars{}{0.779} & \pstars{-}{{---}} & \pstars{-}{{---}} & \pstars{-}{{---}} \\
\hline
\multirow[c]{3}{*}{\makebox[6pt][l]{\rotatebox[origin=c]{90}{other}}} & GINI & \pstars{-}{{---}} & {\cellcolor[HTML]{F3F8FE}} \color[HTML]{000000} \pstars{***}{0.664} & \pstars{-}{{---}} & {\cellcolor[HTML]{083674}} \color[HTML]{F1F1F1} \pstars{***}{0.633} & \pstars{-}{{---}} \\
 & Wikipedia & {\cellcolor[HTML]{A6CEE4}} \color[HTML]{000000} \pstars{***}{0.334} & {\cellcolor[HTML]{F7FBFF}} \color[HTML]{000000} \pstars{***}{0.661} & {\cellcolor[HTML]{F7FBFF}} \color[HTML]{000000} \pstars{***}{0.455} & {\cellcolor[HTML]{7AB6D9}} \color[HTML]{000000} \pstars{***}{0.466} & {\cellcolor[HTML]{F7FBFF}} \color[HTML]{000000} \pstars{***}{0.329} \\
 & wordfreq & {\cellcolor[HTML]{F7FBFF}} \color[HTML]{000000} \pstars{***}{0.242} & {\cellcolor[HTML]{08468B}} \color[HTML]{F1F1F1} \pstars{**}{0.771} & {\cellcolor[HTML]{08306B}} \color[HTML]{F1F1F1} \pstars{}{\mathbf{0.632}} & {\cellcolor[HTML]{3F8FC5}} \color[HTML]{F1F1F1} \pstars{***}{0.522} & {\cellcolor[HTML]{2373B6}} \color[HTML]{F1F1F1} \pstars{***}{0.495} \\
\hline
\multirow[c]{4}{*}{\makebox[6pt][l]{\rotatebox[origin=c]{90}{our\vphantom{l}}}} & TUBELEX\textsubscript{default} & {\cellcolor[HTML]{08306B}} \color[HTML]{F1F1F1} \pstars{-}{\mathbf{0.506}} & {\cellcolor[HTML]{083877}} \color[HTML]{F1F1F1} \pstars{-}{0.777} & {\cellcolor[HTML]{083A7A}} \color[HTML]{F1F1F1} \pstars{-}{0.625} & {\cellcolor[HTML]{083D7F}} \color[HTML]{F1F1F1} \pstars{-}{0.624} & {\cellcolor[HTML]{083776}} \color[HTML]{F1F1F1} \pstars{-}{0.547} \\
 & TUBELEX\textsubscript{regex} & \pstars{-}{{---}} & {\cellcolor[HTML]{083877}} \color[HTML]{F1F1F1} \pstars{}{0.777} & {\cellcolor[HTML]{08478D}} \color[HTML]{F1F1F1} \pstars{**}{0.617} & \pstars{-}{{---}} & {\cellcolor[HTML]{083877}} \color[HTML]{F1F1F1} \pstars{}{0.545} \\
 & TUBELEX\textsubscript{base} & \pstars{-}{{---}} & \pstars{-}{{---}} & \pstars{-}{{---}} & {\cellcolor[HTML]{08306B}} \color[HTML]{F1F1F1} \pstars{***}{\mathbf{0.641}} & \pstars{-}{{---}} \\
 & TUBELEX\textsubscript{lemma} & \pstars{-}{{---}} & {\cellcolor[HTML]{084082}} \color[HTML]{F1F1F1} \pstars{}{0.774} & {\cellcolor[HTML]{08458A}} \color[HTML]{F1F1F1} \pstars{}{0.618} & {\cellcolor[HTML]{08326E}} \color[HTML]{F1F1F1} \pstars{***}{0.637} & {\cellcolor[HTML]{08326E}} \color[HTML]{F1F1F1} \pstars{}{0.551} \\
\bottomrule
\end{tabular}

    \caption{Word familiarity correlation. Strongest (highest) correlations for each language are in bold.
    }
    \label{tab:fam}
\end{table*}

The results in \autoref{tab:ldt} show that in each of the three languages OpenSubtitles and TUBELEX are among the top similarly performing corpora. While OpenSubtitles achieve a stronger correlation for English, \Tdefault achieves a stronger correlation for Chinese. %

Other corpora either performed comparably, but only covered a single language (SubIMDB, EsPal%
), or underperformed noticeably in at least one language (SUBTLEX in Spanish, %
wordfreq in Chinese, Wikipedia in all languages, and GINI and ACTIV-ES in the one language they cover). Furthermore, \autoref{fig:ldt-size} shows that TUBELEX and SUBTLEX perform remarkably well relative to their size.

\subsection{Word Familiarity}
\label{sec:fam}

Word familiarity is a subjective rating of exposure to a given word. Among the subjective variables measured for words in psycholinguistics, it is typically the one most strongly correlated with frequency, and norms for it are available for a wide array of languages.

We measure correlation (PCC) with mean word familiarity from five databases:
Chinese familiarity ratings \citep{su_etal_2023_familiarity},
English MRC lexical database \citep{coltheart_1981_mrc,coltheart_wilson_1987_mrc}, 
Indonesian lexical norms \citep{sianipar_etal_2016_affective},
Japanese word familiarity ratings \citep{asahara-2019-word}\footnote{
    We use the published ratings for reception estimated using a Bayesian linear mixed model.
},
and Spanish lexical norms \citep{guasch_etal_2016_spanish}.
Evaluation on three alternative, smaller databases for English, Spanish, and Japanese can be found in \autoref{sec:alt-fam}.

As shown in \autoref{tab:fam}, in Japanese, \Tbase's correlation is the strongest one, and in all other languages \Tdefault's correlation is either the strongest one or not significantly weaker. Correlations without any significant difference from \Tdefault are achieved by SUBTLEX in Chinese, by all subtitle corpora in English, by wordfreq in Indonesian, and by all subtitle corpora except EsPal, and by CREA-Spoken in Spanish. In Japanese, GINI achieves a remarkably strong correlation, significantly stronger than \Tdefault but still significantly weaker than \Tlemma ($^{\ast\ast}$) and \Tbase ($^{\ast\ast\ast}$).\footnote{We computed the levels of significance separately, as \autoref{tab:fam} compares all corpora only against \Tdefault.}

\begin{figure}[t]
    \centering
    \includegraphics[width=1.0\linewidth]{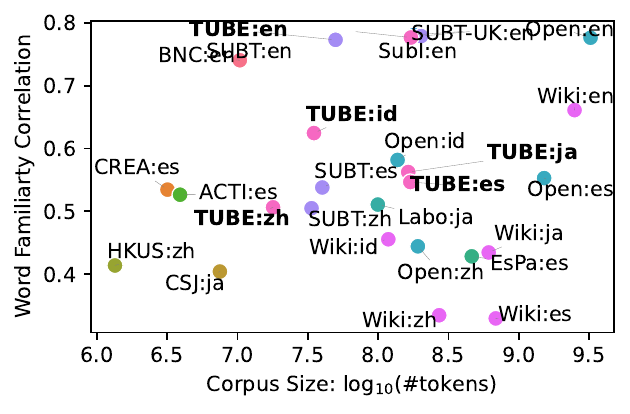}
    \caption{Word familiarity correlation and corpus size. Labeled ``\textit{corpus abbr.}:\textit{lang.\ code}'', ``TUBE'' is \Tdefault, not showing outlier ``Open:ja''.}
    \label{fig:fam-size}
\end{figure}
No corpus achieves results comparable to TUBELEX results across all five languages, but SUBTLEX corpora do not differ significantly on the three languages where they are available. Similarly to LDT, \autoref{fig:fam-size} shows that TUBELEX and SUBTLEX perform remarkably well relative to their size, roughly one order of magnitude smaller than the OpenSubtitles corpora.

\subsection{Lexical Complexity}
\label{sec:lcp}

\begin{table}[t]
    \setlength\tabcolsep{3.5pt}
    \footnotesize
    \centering
    \begin{tabular}{llccc}
\toprule
{} & {Corpus} & {English} & {Japanese} & {Spanish} \\
\midrule
\multirow[c]{3}{*}{\makebox[6pt][l]{\rotatebox[origin=c]{90}{speech}}} & BNC-Spoken & {\cellcolor[HTML]{0F5AA3}} \color[HTML]{F1F1F1} \pstars{***}{-0.695} & \pstars{-}{{---}} & \pstars{-}{{---}} \\
 & CREA-Spoken & \pstars{-}{{---}} & \pstars{-}{{---}} & {\cellcolor[HTML]{BED8EC}} \color[HTML]{000000} \pstars{***}{-0.508} \\
 & CSJ & \pstars{-}{{---}} & {\cellcolor[HTML]{1966AD}} \color[HTML]{F1F1F1} \pstars{***}{-0.563} & \pstars{-}{{---}} \\
\hline
\multirow[c]{7}{*}{\makebox[6pt][l]{\rotatebox[origin=c]{90}{film/TV subtitles}}} & ACTIV-ES & \pstars{-}{{---}} & \pstars{-}{{---}} & {\cellcolor[HTML]{B2D2E8}} \color[HTML]{000000} \pstars{***}{-0.516} \\
 & EsPal & \pstars{-}{{---}} & \pstars{-}{{---}} & {\cellcolor[HTML]{084F99}} \color[HTML]{F1F1F1} \pstars{}{-0.627} \\
 & LaboroTV1+2 & \pstars{-}{{---}} & {\cellcolor[HTML]{084D96}} \color[HTML]{F1F1F1} \pstars{**}{-0.610} & \pstars{-}{{---}} \\
 & OpenSubtitles & {\cellcolor[HTML]{084A91}} \color[HTML]{F1F1F1} \pstars{***}{-0.721} & {\cellcolor[HTML]{F7FBFF}} \color[HTML]{000000} \pstars{***}{-0.191} & {\cellcolor[HTML]{084D96}} \color[HTML]{F1F1F1} \pstars{}{-0.628} \\
 & SubIMDB & {\cellcolor[HTML]{084C95}} \color[HTML]{F1F1F1} \pstars{***}{-0.717} & \pstars{-}{{---}} & \pstars{-}{{---}} \\
 & SUBTLEX & {\cellcolor[HTML]{0E59A2}} \color[HTML]{F1F1F1} \pstars{***}{-0.696} & \pstars{-}{{---}} & {\cellcolor[HTML]{0F5AA3}} \color[HTML]{F1F1F1} \pstars{}{-0.618} \\
 & SUBTLEX-UK & {\cellcolor[HTML]{084990}} \color[HTML]{F1F1F1} \pstars{**}{-0.724} & \pstars{-}{{---}} & \pstars{-}{{---}} \\
\hline
\multirow[c]{3}{*}{\makebox[6pt][l]{\rotatebox[origin=c]{90}{other}}} & GINI & {\cellcolor[HTML]{F7FBFF}} \color[HTML]{000000} \pstars{***}{-0.349} & {\cellcolor[HTML]{94C4DF}} \color[HTML]{000000} \pstars{***}{-0.379} & \pstars{-}{{---}} \\
 & Wikipedia & {\cellcolor[HTML]{2676B8}} \color[HTML]{F1F1F1} \pstars{***}{-0.651} & {\cellcolor[HTML]{4090C5}} \color[HTML]{F1F1F1} \pstars{***}{-0.487} & {\cellcolor[HTML]{F7FBFF}} \color[HTML]{000000} \pstars{***}{-0.454} \\
 & wordfreq & {\cellcolor[HTML]{08306B}} \color[HTML]{F1F1F1} \pstars{}{-0.761} & {\cellcolor[HTML]{084F99}} \color[HTML]{F1F1F1} \pstars{**}{-0.605} & {\cellcolor[HTML]{60A7D2}} \color[HTML]{F1F1F1} \pstars{***}{-0.559} \\
\hline
\multirow[c]{4}{*}{\makebox[6pt][l]{\rotatebox[origin=c]{90}{our\vphantom{l}}}} & TUBELEX\textsubscript{default} & {\cellcolor[HTML]{08306B}} \color[HTML]{F1F1F1} \pstars{-}{\mathbf{-0.762}} & {\cellcolor[HTML]{08306B}} \color[HTML]{F1F1F1} \pstars{-}{\mathbf{-0.661}} & {\cellcolor[HTML]{1E6DB2}} \color[HTML]{F1F1F1} \pstars{-}{-0.604} \\
 & TUBELEX\textsubscript{regex} & {\cellcolor[HTML]{08306B}} \color[HTML]{F1F1F1} \pstars{**}{-0.761} & \pstars{-}{{---}} & {\cellcolor[HTML]{3383BE}} \color[HTML]{F1F1F1} \pstars{*}{-0.588} \\
 & TUBELEX\textsubscript{base} & \pstars{-}{{---}} & {\cellcolor[HTML]{08316D}} \color[HTML]{F1F1F1} \pstars{}{-0.658} & \pstars{-}{{---}} \\
 & TUBELEX\textsubscript{lemma} & {\cellcolor[HTML]{083877}} \color[HTML]{F1F1F1} \pstars{}{-0.749} & {\cellcolor[HTML]{08468B}} \color[HTML]{F1F1F1} \pstars{**}{-0.622} & {\cellcolor[HTML]{08306B}} \color[HTML]{F1F1F1} \pstars{**}{\mathbf{-0.650}} \\
\bottomrule
\end{tabular}

    \caption{Lexical complexity correlation. Strongest (lowest) correlations for each language are in bold.}
    \label{tab:lcp}
\end{table}

\begin{figure}[t]
    \centering
    \includegraphics[width=1.0\linewidth]{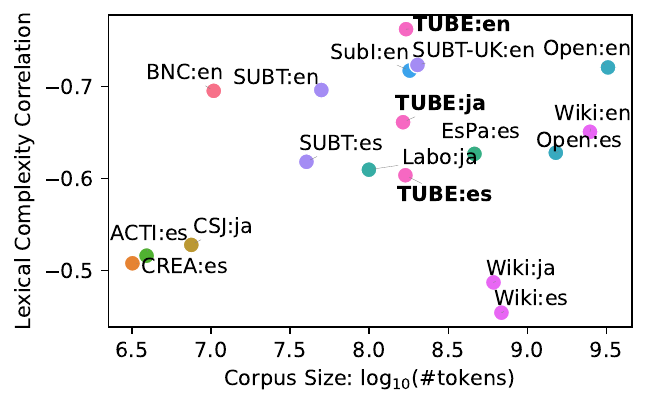}
    \caption{Lexical complexity correlation and corpus size. Labeled ``\textit{corpus abbr.}:\textit{lang.\ code}'', ``TUBE'' is \Tdefault, not showing outlier ``Open:ja''.}
    \label{fig:lcp-size}
\end{figure}

\begin{table}[t]
    \setlength\tabcolsep{5pt}%
    \footnotesize
    \centering
    \begin{tabular}{llS[table-format=-1.3]S[table-format=-1.3]S[table-format=-1.3]}
\toprule
{} & {Corpus / ST System} & {English} & {Japanese} & {Spanish} \\
\midrule
\multirow[c]{3}{*}{\makebox[6pt][l]{\rotatebox[origin=c]{90}{speech}}} & BNC-Spoken & {\cellcolor[HTML]{1561A9}} \color[HTML]{F1F1F1} 0.475 & {---} & {---} \\
 & CREA-Spoken & {---} & {---} & {\cellcolor[HTML]{4E9ACB}} \color[HTML]{F1F1F1} 0.186 \\
 & CSJ & {---} & {\cellcolor[HTML]{1C6AB0}} \color[HTML]{F1F1F1} 0.306 & {---} \\
\hline
\multirow[c]{7}{*}{\makebox[6pt][l]{\rotatebox[origin=c]{90}{film/TV subtitles}}} & ACTIV-ES & {---} & {---} & {\cellcolor[HTML]{F7FBFF}} \color[HTML]{000000} -0.253 \\
 & EsPal & {---} & {---} & {\cellcolor[HTML]{549FCD}} \color[HTML]{F1F1F1} 0.170 \\
 & LaboroTV1+2 & {---} & {\cellcolor[HTML]{0B559F}} \color[HTML]{F1F1F1} 0.349 & {---} \\
 & OpenSubtitles & {\cellcolor[HTML]{2070B4}} \color[HTML]{F1F1F1} 0.445 & {\cellcolor[HTML]{CBDEF1}} \color[HTML]{000000} 0.019 & {\cellcolor[HTML]{1A68AE}} \color[HTML]{F1F1F1} 0.332 \\
 & SubIMDB & {\cellcolor[HTML]{4191C6}} \color[HTML]{F1F1F1} 0.377 & {---} & {---} \\
 & SUBTLEX & {\cellcolor[HTML]{3989C1}} \color[HTML]{F1F1F1} 0.394 & {---} & {\cellcolor[HTML]{F7FBFF}} \color[HTML]{000000} -0.254 \\
 & SUBTLEX-UK & {\cellcolor[HTML]{084F99}} \color[HTML]{F1F1F1} 0.513 & {---} & {---} \\
\hline
\multirow[c]{3}{*}{\makebox[6pt][l]{\rotatebox[origin=c]{90}{other}}} & GINI & {\cellcolor[HTML]{F7FBFF}} \color[HTML]{000000} 0.041 & {\cellcolor[HTML]{99C7E0}} \color[HTML]{000000} 0.105 & {---} \\
 & Wikipedia & {\cellcolor[HTML]{4997C9}} \color[HTML]{F1F1F1} 0.365 & {\cellcolor[HTML]{4090C5}} \color[HTML]{F1F1F1} 0.231 & {\cellcolor[HTML]{F7FBFF}} \color[HTML]{000000} -0.255 \\
 & wordfreq & {\cellcolor[HTML]{08306B}} \color[HTML]{F1F1F1} \bfseries 0.578 & {\cellcolor[HTML]{084E98}} \color[HTML]{F1F1F1} 0.364 & {\cellcolor[HTML]{2F7FBC}} \color[HTML]{F1F1F1} 0.268 \\
\hline
\multirow[c]{4}{*}{\makebox[6pt][l]{\rotatebox[origin=c]{90}{our\vphantom{l}}}} & TUBELEX\textsubscript{default} & {\cellcolor[HTML]{083B7C}} \color[HTML]{F1F1F1} 0.553 & {\cellcolor[HTML]{083979}} \color[HTML]{F1F1F1} 0.405 & {\cellcolor[HTML]{1B69AF}} \color[HTML]{F1F1F1} 0.328 \\
 & TUBELEX\textsubscript{regex} & {\cellcolor[HTML]{083C7D}} \color[HTML]{F1F1F1} 0.552 & {---} & {\cellcolor[HTML]{2070B4}} \color[HTML]{F1F1F1} 0.308 \\
 & TUBELEX\textsubscript{base} & {---} & {\cellcolor[HTML]{08306B}} \color[HTML]{F1F1F1} \bfseries 0.424 & {---} \\
 & TUBELEX\textsubscript{lemma} & {\cellcolor[HTML]{083776}} \color[HTML]{F1F1F1} 0.561 & {\cellcolor[HTML]{3F8FC5}} \color[HTML]{F1F1F1} 0.234 & {\cellcolor[HTML]{2373B6}} \color[HTML]{F1F1F1} 0.299 \\
\hline
\multirow[c]{3}{*}{\makebox[6pt][l]{\rotatebox[origin=c]{90}{top ST}}} & Archaelogy (ID=2) & {\cellcolor[HTML]{2272B6}} \color[HTML]{F1F1F1} 0.439 & {\cellcolor[HTML]{F7FBFF}} \color[HTML]{000000} -0.098 & {\cellcolor[HTML]{3C8CC3}} \color[HTML]{F1F1F1} 0.230 \\
 & GMU (ID=1) & {\cellcolor[HTML]{084A91}} \color[HTML]{F1F1F1} 0.525 & {\cellcolor[HTML]{E1EDF8}} \color[HTML]{000000} -0.039 & {\cellcolor[HTML]{C7DCEF}} \color[HTML]{000000} -0.073 \\
 & TMU-HIT (ID=2) & {\cellcolor[HTML]{084E98}} \color[HTML]{F1F1F1} 0.515 & {\cellcolor[HTML]{083573}} \color[HTML]{F1F1F1} 0.413 & {\cellcolor[HTML]{08306B}} \color[HTML]{F1F1F1} \bfseries 0.494 \\
\bottomrule
\end{tabular}

    \caption{Coefficient of determination $R^2$ achieved in lexical complexity prediction, compared with top shared task systems (top ST), citing their results from \citet{shardlow-etal-2024-bea}. Best (highest) results for each language are in bold.}
    \label{tab:lcp-r2}
\end{table}

Lexical complexity is a subjective rating of word comprehension difficulty in a sentence context. Its prediction can be used for the practical NLP task of lexical simplification. The MultiLS dataset \citep{shardlow-etal-2024-bea}, which we use for evaluation, was annotated by non-native speakers (Japanese) or a mix of natives and non-natives (English and Spanish), whereas the two previously evaluated psycholinguistic tasks only use data collected from natives.
No lexical complexity dataset is available for Chinese or Indonesian. %

As shown in \autoref{tab:lcp}, the strongest correlation in English and Japanese is achieved by \Tdefault, and in Spanish by \Tlemma. Correlations without any significant difference from \Tdefault are achieved by wordfreq in English, and all subtitle corpora except ACTIV-ES for Spanish. Similarly to the previous tasks, \autoref{fig:lcp-size} shows that TUBELEX and SUBTLEX perform remarkably well relative to their size.

As the dataset was used for evaluation of lexical complexity prediction in a shared task \citep{shardlow-etal-2024-bea}, we also compare predictions based on TUBELEX with top shared task participants. To do so, we fit a linear regression model using a single variable, log-frequency or GINI values to the shared task's trial data (30 instances for each language), and clip the predicted values to the range [0, 1]. We compare our results with the shared task submissions that achieved the highest coefficient of determination $R^2$ (and also the highest correlation) in individual languages (TMU-HIT, \citealp{enomoto-etal-2024-tmu}; and GMU, \citealp{goswami-etal-2024-gmu}), and on average across all shared task languages (Archaeology, \citealp{cristea-nisioi-2024-machine}).

As shown in \autoref{tab:lcp-r2}, the best results are achieved by \Tlemma, closely followed by \Tdefault, in English; \Tbase, closely followed by TMU-HIT and \Tdefault, in Japanese; and by TMU-HIT in Spanish, where it outperformed others by a large margin. As we are using cited $R^2$ values achieved by task participants, we cannot evaluate statistical significance in this case. %

A simple linear regression using TUBELEX frequencies has therefore outperformed the top shared task submissions in English and Japanese, namely gradient boosting using multiple features (Archaelogy), ensemble of finetuned BERT models (GMU), and GPT-4 few-shot chain-of-thought prompting (TMU-HIT). It also outperformed the first two of them in Spanish.

It should be noted that, in this case, we are evaluating the prediction of lexical complexity using $R^2$, not a mere correlation with lexical complexity.\footnote{
    While $R^2$ can also be defined as the square of PCC, we use the definition consistent with the shared task evaluation \citep{shardlow-etal-2024-bea}, and implemented in scikit-learn (\url{https://scikit-learn.org/}) as \texttt{r2\_score}. With this definition, $R^2$ is a linear function of the mean squared error, therefore penalizing misprediction of mean and variance. 
}

If we looked only at correlation, thus ignoring misprediction of mean and variance, TMU-HIT's predictions would be more strongly correlated than TUBELEX's in the three languages, and those of the other two systems in English (complete results provided in \autoref{sec:lcp-corr}). This may indicate limitations of LLM prompting as a regression method.

\section{Discussion}
\label{sec:discussion}

\subsection{Tokenization and Lemmatization}

The differences between TUBELEX variants in the evaluation were generally small, but a few observations can be made about each language:

For English, the default variant performs the best across the tasks, but the simpler regex tokenization is never significantly worse. Both always outperform lemmatization. %

In the evaluation of Indonesian, limited to familiarity, default tokenization performed the best.

For Japanese, the base form performs the best for familiarity, and default tokenization for non-native lexical complexity. Both always outperform the orthographically normalized lemma, which show the importance of the exact written form in Japanese.

For Spanish, lemma performed the best for both familiarity and lexical complexity. Regex and default tokenization performed well in LDT only because the data is already limited to uninflected words. For an inflected language such as Spanish, lemmatization has a clear benefit.

\subsection{Spoken vs. Written Language}
\label{sec:spoken-vs-written}

There is a continuum of what we might call spoken and written language in simple terms. 
During human evaluation we have, for instance, observed that speakers in some videos are reading aloud. On one hand, this would make the subtitles representative of written language, not spoken language. On the other hand, the texts being read cover diverse registers (e.g.\ the Bible, professionally announced news, or a pre-written speech), and the speakers often shift between reading and commenting. %
Singing, recitation, scripted acting, and speech rehearsed to various degrees, all appear on YouTube and fall on this written-spoken continuum.
Perhaps surprisingly, the same issues apply to corpora of ``spontaneous'' speech, as they often collect speeches that are prepared (e.g.\ whole CSJ, news in CREA-Spoken). %

In human evaluation (\autoref{sec:human-eval}) we have labeled two categories in TUBELEX that we think require attention: songs, which could be over-represented on YouTube compared to everyday exposure, and synthesized speech, which is effectively written language in disguise.

Overall, we believe that the diverse content found on YouTube contributes to the representativeness of the whole spectrum of spoken language at the small expense of including some amount of written language or songs. This contrasts with most speech corpora, as well as and film subtitle corpora. Speech corpora typically restrict the type or topics of speech they contain by design.

For instance, CSJ is limited  to prepared monologue in common Japanese \citep{maekawa-etal-2000-spontaneous}, omitting any dialogue and dialect, and HKUST/MTS \citep{liu_etal_2006_hkust} is limited to dialogue about 40 specified topics. Film and TV subtitle corpora, on the other hand, consist predominantly of scripted dialogue.

\subsection{Corpus Size}

Previous studies found that in addition to depending on corpus content, correlation with LDT or familiarity grows approximately logarithmically with corpus size \citep{tanaka-ishii_terada_2011_word,paetzold-specia-2016-collecting}. For corpora over $10^7$ tokens, such growth reflects better frequency estimates for low-frequency words, and is measurable and statistically significant only if such words are sufficiently represented in the evaluation dataset.

As we built TUBELEX using a fixed size (120,000) sample of videos for each language, the final corpus size depends on the number of valid videos after cleaning (see \autoref{tab:stats}). 
The Chinese (18M tokens) and Indonesian (38M tokens) corpora are substantially smaller than the others (163M to 171M tokens).
We therefore expect that improvements in correlation could be made by collecting larger corpora, particularly for these two languages, although the effect could be difficult to assess on available data. Increased corpus size would also likely benefit language models (see \autoref{sec:eval-embed}).

Sizes of all corpora and datasets used in this study can be found in \autoref{sec:stats-corpora} and in \autoref{sec:stats-data}, respectively. %

\subsection{Beyond Frequency}
\label{sec:beyond}

Our goal was to evaluate the TUBELEX corpus as an approximation of spoken vocabulary. For a comprehensive comparison with other corpora, we limited the evaluation to word frequency -- the only statistic available for many of the compared corpora. Similarly, most of the data we used for evaluation (LDT, familiarity and complexity datasets) is limited to or focused on single-word items. There are several ways TUBELEX could be used or extended for other purposes:

\paragraph{Language modeling.} More complex language models would allow further applications of the corpus data such as modeling human surprisal (e.g.\ \citealp{wilcox-etal-2023-testing}). We used the current corpus data to train two basic language models: an $n$-gram model and word embeddings, which achieved a mixed performance in our evaluation (\autoref{sec:eval-embed}). Collecting larger data may improve the performance of the embeddings and it would be essential for training more complex language models (masked language models or generative language models). We expect that, similarly to the results achieved by the current embeddings, models trained exclusively on subtitles would lack in some areas, and consequently that training on mixed data would be suitable for a wide range of applications.

\paragraph{Measures of dispersion and contextual diversity.} Single-word corpus statistics are not limited to frequency. We have compared TUBELEX frequency to GINI, a measure of dispersion based on Twitter data, but we leave evaluation of various dispersion metrics computed from TUBELEX for future research. Computation of dispersion metrics generally requires a corpus divided into suitable units such as documents, which TUBELEX provides (videos and channels). Another metric that uses such units is contextual diversity (the number of units in which a word occurs), proposed by \citet{adelman_etal_2006_contextual} as an alternative to frequency for psycholinguistic modeling. TUBELEX word lists readily provide contextual diversity as numbers of videos and channels for each word.

\paragraph{Combining sources.} Combination of frequencies from multiple corpora often achieve more robust results than using a single corpus. While TUBELEX often outperformed wordfreq, which is a resource combining multiple corpus frequencies, combining TUBELEX frequencies with other sources (e.g.\ film subtitles) could also result in a more robust performance, especially for languages underrepresented on YouTube.

\section{Conclusion}

We built a YouTube subtitle corpus of untranslated Chinese, English, Indonesian, Japanese, and Spanish. The frequencies showed consistently strong correlation with LDT, word familiarity, and lexical complexity across the languages. In a comparison with film and TV subtitle corpora, speech corpora, and other common frequency resources, only the SUBTLEX corpora were comparable in correlation strength and consistence. TUBELEX, however, covers Japanese and Indonesian, for which a SUBTLEX corpus is not available. TUBELEX also excelled in the practical task of lexical complexity prediction, where a linear regression based on its frequencies not only outperformed all subtitle and speech corpora but also all submissions in a recent shared task on English and Japanese and all but one on Spanish.

TUBELEX data can be easily used in applications that require spoken vocabulary frequencies (examples given in \autoref{sec:eval-app}), which have typically relied on film subtitles but lacked a suitable resource for many languages. Our method can be used to create TUBELEX corpora for additional languages as well. We see extension beyond using single-word frequencies (outlined in \autoref{sec:beyond}) as a promising direction for future research.

\section*{Limitations}

We focused on evaluation of our corpus in terms of unigram frequencies as an approximation of language exposure, evaluating them using psycholinguistic data and lexical complexity. %

While we also released the higher $n$-grams based on our corpus, we did not evaluate them. We provided only limited evaluation of the word embeddings trained on the corpus (in \autoref{sec:eval-embed}). While our embeddings outperformed those based on Wikipedia in the word similarity task, they achieved lower scores than the much larger OpenSubtitles corpus. We assume this to be an effect of modest corpus size, and expect this to affect the $n$-gram model as well. In this work, we artificially limited the subtitles collected from YouTube to a quantity suitable for modeling unigram frequency. In future work, we plan to explore training more complex models from more extensive YouTube data or joint training from subtitle and written data.

While TUBELEX exceeds traditional speech corpora in size and outperforms them in our evaluation, it has serious limitations for linguistic research. Compared to most specialized speech corpora, it lacks information about types of speech or demographic composition, and suffers from varying quality of transcription. TUBELEX is not limited to any particular language standard and mixes both different language varieties and registers.

We have only collected and evaluated data for five languages. By intentionally selecting a diverse set of languages, however, we demonstrated that our approach is widely applicable to languages with a large enough presence on YouTube. We made our complete source code available for others to reuse and extend.

Estimating how much subtitle data for a particular language is available on YouTube requires non-trivial effort. For a reliable estimate, it is necessary to identify videos by keywords search, scraping the metadata for each video, and evaluating at least a sample of the subtitles using automated language identification (see \autoref{sec:construction}). We did this for two additional languages: For Czech, which has much fewer speakers than any of the current five languages (9.6~million L1~speakers; \citealp{eberhard_etal_2024_ethnologue}), we found enough data to build a corpus comparable to the current five TUBELEX language. We could not, however, find enough data for Pashto with 44~million L1~speakers (ibid.). We hypothesize that this reflects not only the number of speakers and the popularity of YouTube among them, but also the relative prestige of Pashto in the two multi-lingual countries where it is mainly spoken, Afghanistan and Pakistan. Similar challenges may also affect other low-resource languages.

\section*{Ethical Considerations}

The content of YouTube subtitles is copyrighted, which precludes us from distributing the full text of the corpus. In terms of copyright, it is no different from film subtitles, but since YouTube consists of user-generated content, we also had to consider the privacy of the video authors.

We only downloaded subtitles for videos that could be found using the YouTube website search function. The search function is restricted to public videos, excluding any unlisted or private videos.\footnote{
\url{https://support.google.com/youtube/answer/157177?hl=en}
}

None of the data that we have released contains identification of individual videos, channels, or video uploaders. %
The statistical language models we have released contain sequences of at most five consecutive words. The released data does not contain longer excerpts from the original subtitles.

We anonymized multiple kinds of potentially sensitive information before we derived any frequency lists or models from it. In particular, we masked email addresses, HTTP(S) URLs, apparent web URLs without an explicit protocol (e.g.\ \texttt{x.com/username}), apparent social network handles starting with \texttt{@}, and all sequences of digits, which are the primary constituent of phone numbers, IP addresses, account numbers, and other personally identifying information.

As we have accessed YouTube without sign-in, our corpus does not contain any subtitles for age-restricted videos, which YouTube defines as not appropriate for viewers under 18.\footnote{
    \url{https://support.google.com/youtube/answer/2802167?hl=en}
} Note that while YouTube's age restriction also applies to ``excessive profanity'', some subtitles in our corpus still contain vulgar or otherwise inappropriate language, which we did not attempt to remove.

\bibliography{anthology,tubelex_custom} %

\clearpage
\appendix
\onecolumn

\clearpage

\section{Preprocessing and Hyperparameters for Word Embeddings and Statistical Language Models}
\label{sec:model-settings}

\begin{table*}[h]
    \centering
    \setlength\tabcolsep{4pt}
\begin{tabular}{l>{\raggedright\arraybackslash}p{0.78\linewidth}}
\toprule
\multicolumn{2}{c}{
    \raisebox{-0.5ex}{\textbf{Preprocessing}}
    }\\[1ex]
\midrule
Sentence Splitting & On subtitle cue boundaries, and rule-based using PySBD 0.3.4 \citep{sadvilkar_neumann_2020_pysbd}, with rules added for Indonesian based on InaNLP \citep{purwarianti_etal_2016_inanlp}.\\
\midrule
Tokenization & TUBELEX regex tokenization for English, Japanese, and Spanish, and default tokenization for Japanese and Spanish (details in \autoref{sec:tok-freq-lists}).\\
\midrule
Normalization & Lower case, Unicode NFKC \citep{whistler_2023_uax}.\\
\midrule
\multicolumn{2}{c}{
    \raisebox{-0.5ex}{\textbf{Hyperparameters}}
    }\\[1ex]
\midrule

Word Embeddings &
300-dimensional fastText CBOW model with position weights, 10 negative samples, 10~epochs, character 5-grams, other: default \citep{grave-etal-2018-learning}.\newline
\TableItem{software: \url{https://github.com/facebookresearch/fastText}}
\TableItem{CLI: \texttt{fasttext cbow -dim 300 -neg 10 -epoch 10 -minn 5 -maxn 5}}\\
\midrule

Statistical Model & Modified Kneser-Ney language model of order 5 \citep{heafield-etal-2013-scalable}.\newline
\TableItem{software: \url{https://kheafield.com/code/kenlm/}}
\TableItem{CLI: \texttt{lmplz -o 5}}\\

\bottomrule

\end{tabular}
\caption{Preprocessing and hyperparameters used to train word embeddings and statistical language models on TUBELEX. We used the same preprocessing for both.}
\end{table*}

\vfill

\section{Sizes of Word Embeddings and Statistical Language Models}
\label{sec:model-sizes}

\begin{table*}[h]
    \centering
\begin{tabular}{l!{\hspace{1em}}rrrrr>{\hspace{1em}}c}
\toprule
\multirow{2}{*}{Language}&
\multicolumn{5}{c}{Statistical Language Model $n$-Grams} &
\multirow{2}{*}{\parbox{\widthof{Vocabulary}}{\centering FastText\\Vocabulary}}\\
\cmidrule{2-6}
&	$n=1$ &	$n=2$ &	$n=3$ &	$n=4$ &	$n=5$ & \\
\midrule
Chinese & 432,670 & 5,760,278 & 12,642,939 & 15,320,475 & 15,264,065 & 114,237\\
English & 420,583 & 12,798,615 & 53,560,054 & 99,046,960 & 126,309,472 & 131,757\\
Indonesian & 300,647 & 6,746,497 & 20,766,383 & 29,052,968 & 31,555,431 & \phantom{0}81,801\\
Japanese & 405,676 & 10,898,295 & 45,793,067 & 85,956,149 & 113,644,170 & 145,429\\
Spanish & 613,056 & 15,482,447 & 62,465,043 & 113,520,573 & 139,641,687 & 197,107\\\bottomrule
\end{tabular}
\caption{Numbers of $n$-grams of the statistical language models  (KenLM) and vocabulary size of the word embeddings (fastText) trained on TUBELEX. (Minimum frequency for the fastText model is 5.)}
\end{table*}

\vfill

\clearpage
\section{Evaluation of Word Embeddings}
\label{sec:eval-embed}

In Tables~\ref{tab:ana} and \ref{tab:sim}, we compare the performance of TUBELEX embeddings in word analogy and word similarity with previously published embeddings trained on Wikipedia \citep{grave-etal-2018-learning}, and OpenSubtitles2018 \citep{vanparidon_thompson_2021_subs2vec}. 

The TUBELEX embeddings were trained using the same hyperparameters (see \autoref{sec:model-settings}) as the Wikipedia embeddings by \citet{grave-etal-2018-learning}, while \citet{vanparidon_thompson_2021_subs2vec} used a different setup. All compared embeddings are fastText, but we do not use character $n$-grams to embed out-of-vocabulary words. For fairness, we always evaluate the whole dataset including out-of-vocabulary words. We only evaluate on English and Spanish, for which comparable evaluation data and pre-trained OpenSubtitles2018 embeddings are available.

In word analogy, TUBELEX embeddings underperform the other embeddings. In word similarity, they outperform Wikipedia, but slightly underperform OpenSubtitles. The overall performance is very close to the OpenSubtitles embeddings, and we hypothesize that the gap between the two is caused by the TUBELEX corpus being an order of magnitude smaller than OpenSubtitles. While we have observed that TUBELEX's size does not affect the quality of unigram frequencies, the word embeddings would likely benefit from a larger corpus.

\vspace{\baselineskip}

\begin{table}[h]
    \small
    \centering
    \begin{tabular}{llccccc}
\toprule
Language & Embeddings & Sem.: Geography & Sem.: Family & Semantic & Syntactic & Total \\
\midrule
\multirow[c]{3}{*}{English} & Wikipedia & 0.775 & 0.822 & 0.778 & 0.721 & 0.747 \\
 & OpenSubtitles & 0.144 & 0.852 & 0.184 & 0.757 & 0.497 \\
 & TUBELEX & 0.142 & 0.626 & 0.170 & 0.628 & 0.420 \\
\hline
\multirow[c]{3}{*}{Spanish} & Wikipedia & 0.466 & 0.863 & 0.484 & 0.572 & 0.524 \\
 & OpenSubtitles & 0.087 & 0.892 & 0.125 & 0.516 & 0.301 \\
 & TUBELEX & 0.064 & 0.839 & 0.101 & 0.501 & 0.281 \\
\bottomrule
\end{tabular}

    \caption{Accuracy in word analogy evaluated on English data
    \citep{mikolov_etal_2013_efficient} and Spanish data derived from it (\url{https://crscardellino.net/SBWCE/}). We list separately accuracy in Geography and Family subcategories of the Semantic category.
    } 
    \label{tab:ana}
\end{table}

\begin{table}[h]
    \small
    \centering
    \begin{tabular}{llcc}
\toprule
Language & Embeddings & Pearson's $r$ & Spearman's $\rho$ \\
\midrule
\multirow[c]{3}{*}{English} & Wikipedia & 0.379 & 0.434 \\
 & OpenSubtitles & 0.468 & 0.532 \\
 & TUBELEX & 0.385 & 0.457 \\
\hline
\multirow[c]{3}{*}{Spanish} & Wikipedia & 0.342 & 0.387 \\
 & OpenSubtitles & 0.445 & 0.475 \\
 & TUBELEX & 0.415 & 0.450 \\
\bottomrule
\end{tabular}

    \caption{Correlation in word similarity evaluated on parallel English and Spanish data from Multi-SimLex \citep{vulic_etal_2021_multi}.} 
    \label{tab:sim}
\end{table}

\section{GINI Metric}
\label{sec:gini}

The GINI metric was proposed by \citet{murayama_etal_2018_word} for simplification and readability assessment. It is inspired by the Gini index of income inequality, and measures words' dispersion in frequency of use by different people. Similarly to TUBELEX, it uses user-generated data (from Twitter), and its pre-computed values are available for English and Japanese. As the original study is available only in Japanese, we summarize its computation for the reader's convenience:

\begin{enumerate}
\item Construct a matrix of numbers of word occurrences shaped $(m, n) =$ (\#users, \#words).
\item Normalize twice: first making each row sum to 1, then making each column sum to 1.
\item For each column (word) $\mathbf{x}$, compute: $Gini(\mathbf{x}) = \frac{1}{2\mu_{\mathbf{x}} n^2} \sum_{i=1}^n \sum_{j=1}^n |x_i-x_j|$
\item To compute the final values of the metric, apply: $-\log(1-Gini(\mathbf{x}))$.
\end{enumerate}

\clearpage

\section{Details of the Evaluated Corpora}
\label{sec:corpora-details}

\autoref{tab:corpora-details} shows details of the evaluated corpora. We compute frequency with Laplace smoothing (see \autoref{sec:frequency}) from them with the following exception: for ACTIV-ES and wordfreq, which do not provide token counts, we instead assign the corpus minimum frequency to missing words. We also directly use GINI values, analogously assigning the corpus maximum value to missing words, as high GINI values indicate high dispersion.
The words featured in our experiments may be tokenized as multiple tokens. We assign them the minimum of the frequencies of the individual tokens, and maximum of the GINI values.
In all experiments, we use the logarithm of frequency. For GINI, we use the additive inverse ($\log(1-Gini(\mathbf{x}))$) of the original metric's values described in \autoref{sec:gini}, for the purpose of comparison with log-frequencies used for the other corpora.

\vspace{\baselineskip}
\noindent%
\begin{minipage}{\textwidth}%
\captionsetup{type=table}%
    \setlength\tabcolsep{4pt}
    \footnotesize
    \centering
\begin{tabular}{
    >{\bfseries}l
    >{\bfseries}p{0.123\linewidth}
    p{0.388\linewidth}
    >{\raggedright\arraybackslash}p{0.403\linewidth}
    }

\toprule

& Corpus & \textbf{Details} & \textbf{Source} \\
\midrule

\TableGroup{10}{speech} 
&
BNC-Spoken &
We construct a frequency list for the spoken subset of BNC by computing a difference of the ``all'' and ``written'' unlemmatized BNC frequency lists compiled by Adam Kilgarif. &
\TableItem{\url{https://www.kilgarriff.co.uk/bnc-readme.html}}\\

&
CREA-Spoken\kern -2em%
&
We use the frequency list of the spoken subset of CREA. &
\citet{alonso_etal_2011_oral}
\\

&
CSJ & We use the published CSJ frequency list lemmatized using MeCab/Unidic. & \citetaliasyear{ninjal_2018_wordlist}
\\

&
HKUST/MTS & We construct a frequency list from the corpus transcripts using the jieba tokenization. &
\TableItem[2pt]{\url{https://catalog.ldc.upenn.edu/LDC2005T32}}
\\

\specialrule{\arrayrulewidth}{\aboverulesep}{\belowrulesep}

\TableGroup{33}{film/TV subtitles} &
ACTIV-ES & We use the published 1-gram frequency list, version 0.2. & \TableItem{\url{https://github.com/francojc/activ-es}}\\

&
EsPal &
We use the public web form, to retrieve frequencies of all tokens for our experiments. & \TableItem{\url{https://www.bcbl.eu/databases/espal/wordidx.php}}
\\

&
LaboroTV1+2 &
LaboroTVSpeech (2020) and LaboroTVSpeech2 (2024): We combine pre-tokenized training and development data of the two releases of to generate a single frequency list. &
\TableItem{\url{https://laboro.ai/activity/column/engineer/eg-laboro-tv-corpus-jp/}}
\TableItem{\url{https://laboro.ai/activity/column/engineer/laborotvspeech2/}}\\

&
OpenSubtitles &
We use the published frequency lists from the updated 2018 version of the collection. For Chinese we use the list identified as ``China mainland'', which mostly uses simplified Chinese characters. 
(In the 2018 version, Chinese subtitles are divided into ``China mainland'' and ``Taiwan''. Details of the division are not documented. The OpenSubtitles website itself divides Chinese into simplified, traditional, and Cantonese.)
&
\TableItem{\url{https://opus.nlpl.eu/OpenSubtitles/&/v2018/OpenSubtitles}}\\

&
SubIMDB &
We generate a frequency list from the full SubIMDB corpus, which comes in a pre-tokenized form. No frequency list was published for the corpus. &
\TableItem{\url{https://zenodo.org/records/2552407}}\\

&
SUBTLEX &
We use the published SUBTLEX raw frequency counts for English (US), Spanish (ESP), and Chinese (CH). & \TableItem{\url{https://www.ugent.be/pp/experimentele-psychologie/en/research/documents/subtlexus/subtlexus2.zip} (US)}
\TableItem{\url{http://www.ugent.be/pp/experimentele-psychologie/en/research/documents/subtlexch/subtlexchwf.zip} (CH)}
\TableItem{\url{https://web.archive.org/web/20220702151524/http://crr.ugent.be/papers/SUBTLEX-ESP.zip}~(ESP)}\\

&
SUBTLEX-UK\kern -2em%
&
For English, we use raw frequency counts from SUBTLEX-UK as well. &
\TableItem{\url{https://www.psychology.nottingham.ac.uk/subtlex-uk/SUBTLEX-UK.txt.zip}}\\

\specialrule{\arrayrulewidth}{\aboverulesep}{\belowrulesep}

\TableGroup{7}{other} &
GINI &
We use the published WORD GINI lists for English and Japanese. (Details about the GINI metric in \autoref{sec:gini}.) &
\TableItem{\url{https://sociocom.naist.jp/word-gini-en/}}\\

&
Wikipedia &
We use frequency lists based on cleaned up Wikipedia text tokenized using a regular expression. &
\TableItem{\url{https://github.com/adno/wikipedia-word-frequency-clean}}
\\

&
wordfreq &
We use the default (large) lists available from the Python library. &
\TableItem{\url{https://pypi.org/project/wordfreq/}}\\

\bottomrule

\end{tabular}
\captionof{table}{Detailed information and sources for the corpora used for evaluation. Source is the publication, if it contains the frequency lists as supplementary material, or an URL from which the data (corpus or frequency list) is available. The corpora are introduced and cited in \autoref{sec:related} and \autoref{sec:eval-corpora}.}
\label{tab:corpora-details}
\vspace{-1\baselineskip} 
\end{minipage}

\clearpage
\section{Statistics of the Evaluated Corpora}
\label{sec:stats-corpora}

\vspace{\baselineskip}
\noindent%
\begin{minipage}{\textwidth}%
\captionsetup{type=table}%
    \setlength\tabcolsep{4pt}
    \centering
    \begin{tabular}{llrrrrr}
\toprule
Corpus &  & Chinese & English & Indonesian & Japanese & Spanish \\
\midrule
\multirow[c]{2}{*}{BNC-Spoken} & tokens & {---} & 10,365,473 & {---} & {---} & {---} \\
 & \cellcolor[HTML]{EEEEEE} types & \cellcolor[HTML]{EEEEEE} {---} & \cellcolor[HTML]{EEEEEE} 669,417 & \cellcolor[HTML]{EEEEEE} {---} & \cellcolor[HTML]{EEEEEE} {---} & \cellcolor[HTML]{EEEEEE} {---} \\
\multirow[c]{2}{*}{CREA-Spoken} & tokens & {---} & {---} & {---} & {---} & 3,171,903 \\
 & \cellcolor[HTML]{EEEEEE} types & \cellcolor[HTML]{EEEEEE} {---} & \cellcolor[HTML]{EEEEEE} {---} & \cellcolor[HTML]{EEEEEE} {---} & \cellcolor[HTML]{EEEEEE} {---} & \cellcolor[HTML]{EEEEEE} 67,979 \\
\multirow[c]{2}{*}{CSJ} & tokens & {---} & {---} & {---} & 7,479,773 & {---} \\
 & \cellcolor[HTML]{EEEEEE} types & \cellcolor[HTML]{EEEEEE} {---} & \cellcolor[HTML]{EEEEEE} {---} & \cellcolor[HTML]{EEEEEE} {---} & \cellcolor[HTML]{EEEEEE} 40,630 & \cellcolor[HTML]{EEEEEE} {---} \\
\multirow[c]{2}{*}{HKUST/MTS} & tokens & 1,342,379 & {---} & {---} & {---} & {---} \\
 & \cellcolor[HTML]{EEEEEE} types & \cellcolor[HTML]{EEEEEE} 42,247 & \cellcolor[HTML]{EEEEEE} {---} & \cellcolor[HTML]{EEEEEE} {---} & \cellcolor[HTML]{EEEEEE} {---} & \cellcolor[HTML]{EEEEEE} {---} \\
\multirow[c]{2}{*}{ACTIV-ES} & tokens & {---} & {---} & {---} & {---} & 3,897,234 \\
 & \cellcolor[HTML]{EEEEEE} types & \cellcolor[HTML]{EEEEEE} {---} & \cellcolor[HTML]{EEEEEE} {---} & \cellcolor[HTML]{EEEEEE} {---} & \cellcolor[HTML]{EEEEEE} {---} & \cellcolor[HTML]{EEEEEE} 80,787 \\
\multirow[c]{2}{*}{EsPal} & tokens & {---} & {---} & {---} & {---} & 462,611,693 \\
 & \cellcolor[HTML]{EEEEEE} types & \cellcolor[HTML]{EEEEEE} {---} & \cellcolor[HTML]{EEEEEE} {---} & \cellcolor[HTML]{EEEEEE} {---} & \cellcolor[HTML]{EEEEEE} {---} & \cellcolor[HTML]{EEEEEE} 35,257 \\
\multirow[c]{2}{*}{LaboroTV1+2} & tokens & {---} & {---} & {---} & 99,367,439 & {---} \\
 & \cellcolor[HTML]{EEEEEE} types & \cellcolor[HTML]{EEEEEE} {---} & \cellcolor[HTML]{EEEEEE} {---} & \cellcolor[HTML]{EEEEEE} {---} & \cellcolor[HTML]{EEEEEE} 218,762 & \cellcolor[HTML]{EEEEEE} {---} \\
\multirow[c]{2}{*}{OpenSubtitles} & tokens & 191,379,324 & 3,235,391,790 & 137,231,876 & 23,665,222 & 1,512,443,143 \\
 & \cellcolor[HTML]{EEEEEE} types & \cellcolor[HTML]{EEEEEE} 1,009,838 & \cellcolor[HTML]{EEEEEE} 2,290,458 & \cellcolor[HTML]{EEEEEE} 456,125 & \cellcolor[HTML]{EEEEEE} 58,856 & \cellcolor[HTML]{EEEEEE} 1,629,907 \\
\multirow[c]{2}{*}{SubIMDB} & tokens & {---} & 179,967,485 & {---} & {---} & {---} \\
 & \cellcolor[HTML]{EEEEEE} types & \cellcolor[HTML]{EEEEEE} {---} & \cellcolor[HTML]{EEEEEE} 899,603 & \cellcolor[HTML]{EEEEEE} {---} & \cellcolor[HTML]{EEEEEE} {---} & \cellcolor[HTML]{EEEEEE} {---} \\
\multirow[c]{2}{*}{SUBTLEX} & tokens & 33,546,516 & 49,719,560 & {---} & {---} & 40,017,237 \\
 & \cellcolor[HTML]{EEEEEE} types & \cellcolor[HTML]{EEEEEE} 99,121 & \cellcolor[HTML]{EEEEEE} 74,286 & \cellcolor[HTML]{EEEEEE} {---} & \cellcolor[HTML]{EEEEEE} {---} & \cellcolor[HTML]{EEEEEE} 94,261 \\
\multirow[c]{2}{*}{SUBTLEX-UK} & tokens & {---} & 201,706,753 & {---} & {---} & {---} \\
 & \cellcolor[HTML]{EEEEEE} types & \cellcolor[HTML]{EEEEEE} {---} & \cellcolor[HTML]{EEEEEE} 160,022 & \cellcolor[HTML]{EEEEEE} {---} & \cellcolor[HTML]{EEEEEE} {---} & \cellcolor[HTML]{EEEEEE} {---} \\
\multirow[c]{2}{*}{GINI} & tokens & {---} & {---} & {---} & {---} & {---} \\
 & \cellcolor[HTML]{EEEEEE} types & \cellcolor[HTML]{EEEEEE} {---} & \cellcolor[HTML]{EEEEEE} 324,713 & \cellcolor[HTML]{EEEEEE} {---} & \cellcolor[HTML]{EEEEEE} 208,275 & \cellcolor[HTML]{EEEEEE} {---} \\
\multirow[c]{2}{*}{Wikipedia} & tokens & 271,230,431 & 2,489,387,103 & 117,956,650 & 610,467,200 & 685,158,870 \\
 & \cellcolor[HTML]{EEEEEE} types & \cellcolor[HTML]{EEEEEE} 1,403,791 & \cellcolor[HTML]{EEEEEE} 2,161,820 & \cellcolor[HTML]{EEEEEE} 373,461 & \cellcolor[HTML]{EEEEEE} 522,210 & \cellcolor[HTML]{EEEEEE} 986,947 \\
\multirow[c]{2}{*}{wordfreq} & tokens & {---} & {---} & {---} & {---} & {---} \\
 & \cellcolor[HTML]{EEEEEE} types & \cellcolor[HTML]{EEEEEE} 334,609 & \cellcolor[HTML]{EEEEEE} 321,180 & \cellcolor[HTML]{EEEEEE} 31,188 & \cellcolor[HTML]{EEEEEE} 214,960 & \cellcolor[HTML]{EEEEEE} 342,072 \\
\multirow[c]{2}{*}{TUBELEX\textsubscript{default}} & tokens & 17,865,686 & 170,750,870 & 34,903,381 & 163,439,781 & 169,188,689 \\
 & \cellcolor[HTML]{EEEEEE} types & \cellcolor[HTML]{EEEEEE} 432,532 & \cellcolor[HTML]{EEEEEE} 467,296 & \cellcolor[HTML]{EEEEEE} 307,633 & \cellcolor[HTML]{EEEEEE} 409,503 & \cellcolor[HTML]{EEEEEE} 632,112 \\
\multirow[c]{2}{*}{TUBELEX\textsubscript{regex}} & tokens & {---} & 170,816,384 & 34,293,878 & {---} & 166,423,254 \\
 & \cellcolor[HTML]{EEEEEE} types & \cellcolor[HTML]{EEEEEE} {---} & \cellcolor[HTML]{EEEEEE} 420,718 & \cellcolor[HTML]{EEEEEE} 300,870 & \cellcolor[HTML]{EEEEEE} {---} & \cellcolor[HTML]{EEEEEE} 613,181 \\
\multirow[c]{2}{*}{TUBELEX\textsubscript{base}} & tokens & {---} & {---} & {---} & 163,439,781 & {---} \\
 & \cellcolor[HTML]{EEEEEE} types & \cellcolor[HTML]{EEEEEE} {---} & \cellcolor[HTML]{EEEEEE} {---} & \cellcolor[HTML]{EEEEEE} {---} & \cellcolor[HTML]{EEEEEE} 378,276 & \cellcolor[HTML]{EEEEEE} {---} \\
\multirow[c]{2}{*}{TUBELEX\textsubscript{lemma}} & tokens & {---} & 170,764,637 & 34,904,605 & 163,462,537 & 169,188,635 \\
 & \cellcolor[HTML]{EEEEEE} types & \cellcolor[HTML]{EEEEEE} {---} & \cellcolor[HTML]{EEEEEE} 433,545 & \cellcolor[HTML]{EEEEEE} 266,827 & \cellcolor[HTML]{EEEEEE} 329,303 & \cellcolor[HTML]{EEEEEE} 527,060 \\
\bottomrule
\end{tabular}

    \captionof{table}{Numbers of tokens and types in the corpora evaluated in \autoref{sec:eval-corpora}. Number of types is always based on the actual frequency lists we use (see \autoref{sec:corpora-details}), after lowercasing and combining equivalent words (a few corpora list separately words differing only in case or POS). Number of tokens are either sums of individual token counts or explicit total token counts if available. GINI and wordfreq data do not report numbers of tokens (only index values and relative frequencies, respectively). Wordfreq also removes types with frequency less than $10^{-8}$.
    }
    \label{tab:stats-corpora}
\end{minipage}

\clearpage

\section{Evaluation on Alternative Word Familiarity Norms}
\label{sec:alt-fam}

\begin{table*}[h]
    \footnotesize
    \centering
    \begin{tabular}{llccc}
\toprule
{} & {} & {English} & {Japanese} & {Spanish} \\
{} & {Corpus} & {(Glasgow)} & {(Amano+Kondo)} & {(Moreno-Martínez)} \\
\midrule
\multirow[c]{3}{*}{\makebox[6pt][l]{\rotatebox[origin=c]{90}{speech}}} & BNC-Spoken & {\cellcolor[HTML]{084387}} \color[HTML]{F1F1F1} \pstars{*}{0.658} & \pstars{-}{{---}} & \pstars{-}{{---}} \\
 & CREA-Spoken & \pstars{-}{{---}} & \pstars{-}{{---}} & {\cellcolor[HTML]{84BCDB}} \color[HTML]{000000} \pstars{***}{0.510} \\
 & CSJ & \pstars{-}{{---}} & {\cellcolor[HTML]{94C4DF}} \color[HTML]{000000} \pstars{***}{0.441} & \pstars{-}{{---}} \\
\hline
\multirow[c]{7}{*}{\makebox[6pt][l]{\rotatebox[origin=c]{90}{film/TV subtitles}}} & ACTIV-ES & \pstars{-}{{---}} & \pstars{-}{{---}} & {\cellcolor[HTML]{A3CCE3}} \color[HTML]{000000} \pstars{***}{0.495} \\
 & EsPal & \pstars{-}{{---}} & \pstars{-}{{---}} & {\cellcolor[HTML]{2F7FBC}} \color[HTML]{F1F1F1} \pstars{**}{0.557} \\
 & LaboroTV1+2 & \pstars{-}{{---}} & {\cellcolor[HTML]{105BA4}} \color[HTML]{F1F1F1} \pstars{***}{0.536} & \pstars{-}{{---}} \\
 & OpenSubtitles & {\cellcolor[HTML]{084D96}} \color[HTML]{F1F1F1} \pstars{}{0.650} & {\cellcolor[HTML]{F7FBFF}} \color[HTML]{000000} \pstars{***}{0.354} & {\cellcolor[HTML]{08306B}} \color[HTML]{F1F1F1} \pstars{}{\mathbf{0.612}} \\
 & SubIMDB & {\cellcolor[HTML]{08306B}} \color[HTML]{F1F1F1} \pstars{***}{\mathbf{0.675}} & \pstars{-}{{---}} & \pstars{-}{{---}} \\
 & SUBTLEX & {\cellcolor[HTML]{0C56A0}} \color[HTML]{F1F1F1} \pstars{}{0.642} & \pstars{-}{{---}} & {\cellcolor[HTML]{0C56A0}} \color[HTML]{F1F1F1} \pstars{}{0.585} \\
 & SUBTLEX-UK & {\cellcolor[HTML]{08316D}} \color[HTML]{F1F1F1} \pstars{***}{0.674} & \pstars{-}{{---}} & \pstars{-}{{---}} \\
\hline
\multirow[c]{3}{*}{\makebox[6pt][l]{\rotatebox[origin=c]{90}{other}}} & GINI & {\cellcolor[HTML]{D9E7F5}} \color[HTML]{000000} \pstars{***}{0.482} & {\cellcolor[HTML]{08306B}} \color[HTML]{F1F1F1} \pstars{***}{\mathbf{0.572}} & \pstars{-}{{---}} \\
 & Wikipedia & {\cellcolor[HTML]{F7FBFF}} \color[HTML]{000000} \pstars{***}{0.446} & {\cellcolor[HTML]{B0D2E7}} \color[HTML]{000000} \pstars{***}{0.423} & {\cellcolor[HTML]{F7FBFF}} \color[HTML]{000000} \pstars{***}{0.430} \\
 & wordfreq & {\cellcolor[HTML]{0F5AA3}} \color[HTML]{F1F1F1} \pstars{**}{0.638} & {\cellcolor[HTML]{2B7BBA}} \color[HTML]{F1F1F1} \pstars{***}{0.510} & {\cellcolor[HTML]{2F7FBC}} \color[HTML]{F1F1F1} \pstars{***}{0.557} \\
\hline
\multirow[c]{4}{*}{\makebox[6pt][l]{\rotatebox[origin=c]{90}{our\vphantom{l}}}} & TUBELEX\textsubscript{default} & {\cellcolor[HTML]{08519C}} \color[HTML]{F1F1F1} \pstars{-}{0.646} & {\cellcolor[HTML]{09529D}} \color[HTML]{F1F1F1} \pstars{-}{0.544} & {\cellcolor[HTML]{08326E}} \color[HTML]{F1F1F1} \pstars{-}{0.610} \\
 & TUBELEX\textsubscript{regex} & {\cellcolor[HTML]{09529D}} \color[HTML]{F1F1F1} \pstars{}{0.646} & \pstars{-}{{---}} & {\cellcolor[HTML]{08326E}} \color[HTML]{F1F1F1} \pstars{}{0.610} \\
 & TUBELEX\textsubscript{base} & \pstars{-}{{---}} & {\cellcolor[HTML]{083979}} \color[HTML]{F1F1F1} \pstars{***}{0.564} & \pstars{-}{{---}} \\
 & TUBELEX\textsubscript{lemma} & {\cellcolor[HTML]{0E59A2}} \color[HTML]{F1F1F1} \pstars{*}{0.639} & {\cellcolor[HTML]{0E59A2}} \color[HTML]{F1F1F1} \pstars{***}{0.538} & {\cellcolor[HTML]{083370}} \color[HTML]{F1F1F1} \pstars{}{0.609} \\
\bottomrule
\end{tabular}

    \caption{Word familiarity (alternative norms) correlation (PCC). Strongest (highest) correlations for each language are in bold. Glasgow norms \citep{scott_etal_2019_glasgow} for English, norms by \citet{moreno-martinez_etal_2014_spanish} for Spanish, and written word familiarity ratings by \citet{amano_kondo_1999_ntt} for Japanese. All three databases are smaller than the ones presented in \autoref{sec:fam}.
    }
    \label{tab:alt-fam}
\end{table*}

\vfill
\section{Evaluation Dataset Sizes}
\label{sec:stats-data}

\begin{table*}[h]
    \centering
    \begin{tabular}{lrrrrr}
\toprule
Task & Chinese & English & Indonesian & Japanese & Spanish \\
\midrule
Lexical Decision Time & 12,576 & 38,130 & {---} & {---} & 45,190 \\
Lexical Complexity & {---} & 570 & {---} & 570 & 593 \\
Word Familiarity & 24,325 & 4,923 & 1,490 & 81,271 & 1,400 \\
Word Familiarity (Alternative) & {---} & 4,682 & {---} & 76,883 & 820 \\
\bottomrule
\end{tabular}

    \caption{Numbers of instances in the datasets used for evaluation. The individual datasets are introduced in \autoref{sec:ldt} for lexical decision time, \autoref{sec:lcp} for lexical complexity, \autoref{sec:fam} for word familiarity, and \autoref{sec:alt-fam} for word familiarity -- alternative datasets.
    }
    \label{tab:stats-data}
\end{table*}

\vfill

\clearpage
\section{Correlation with Lexical Complexity Predictions}
\label{sec:lcp-corr}

\begin{table}[h]
    \footnotesize
    \centering
    \begin{tabular}{llS[table-format=-1.3]S[table-format=-1.3]S[table-format=-1.3]}
\toprule
{} & {Corpus / ST System} & {English} & {Japanese} & {Spanish} \\
\midrule
\multirow[c]{3}{*}{\makebox[6pt][l]{\rotatebox[origin=c]{90}{speech}}} & BNC-Spoken & {\cellcolor[HTML]{2D7DBB}} \color[HTML]{F1F1F1} 0.701 & {---} & {---} \\
 & CREA-Spoken & {---} & {---} & {\cellcolor[HTML]{1460A8}} \color[HTML]{F1F1F1} 0.508 \\
 & CSJ & {---} & {\cellcolor[HTML]{1F6EB3}} \color[HTML]{F1F1F1} 0.565 & {---} \\
\hline
\multirow[c]{7}{*}{\makebox[6pt][l]{\rotatebox[origin=c]{90}{film/TV subtitles}}} & ACTIV-ES & {---} & {---} & {\cellcolor[HTML]{E9F2FA}} \color[HTML]{000000} -0.516 \\
 & EsPal & {---} & {---} & {\cellcolor[HTML]{084A91}} \color[HTML]{F1F1F1} 0.627 \\
 & LaboroTV1+2 & {---} & {\cellcolor[HTML]{125EA6}} \color[HTML]{F1F1F1} 0.610 & {---} \\
 & OpenSubtitles & {\cellcolor[HTML]{2272B6}} \color[HTML]{F1F1F1} 0.721 & {\cellcolor[HTML]{CBDEF1}} \color[HTML]{000000} 0.191 & {\cellcolor[HTML]{084990}} \color[HTML]{F1F1F1} 0.628 \\
 & SubIMDB & {\cellcolor[HTML]{2474B7}} \color[HTML]{F1F1F1} 0.717 & {---} & {---} \\
 & SUBTLEX & {\cellcolor[HTML]{3080BD}} \color[HTML]{F1F1F1} 0.696 & {---} & {\cellcolor[HTML]{F7FBFF}} \color[HTML]{000000} -0.618 \\
 & SUBTLEX-UK & {\cellcolor[HTML]{2070B4}} \color[HTML]{F1F1F1} 0.726 & {---} & {---} \\
\hline
\multirow[c]{3}{*}{\makebox[6pt][l]{\rotatebox[origin=c]{90}{other}}} & GINI & {\cellcolor[HTML]{F7FBFF}} \color[HTML]{000000} 0.349 & {\cellcolor[HTML]{6FB0D7}} \color[HTML]{F1F1F1} 0.379 & {---} \\
 & Wikipedia & {\cellcolor[HTML]{4997C9}} \color[HTML]{F1F1F1} 0.651 & {\cellcolor[HTML]{3C8CC3}} \color[HTML]{F1F1F1} 0.487 & {\cellcolor[HTML]{DFECF7}} \color[HTML]{000000} -0.454 \\
 & wordfreq & {\cellcolor[HTML]{125DA6}} \color[HTML]{F1F1F1} 0.763 & {\cellcolor[HTML]{135FA7}} \color[HTML]{F1F1F1} 0.605 & {\cellcolor[HTML]{0C56A0}} \color[HTML]{F1F1F1} 0.559 \\
\hline
\multirow[c]{4}{*}{\makebox[6pt][l]{\rotatebox[origin=c]{90}{our\vphantom{l}}}} & TUBELEX\textsubscript{default} & {\cellcolor[HTML]{105BA4}} \color[HTML]{F1F1F1} 0.766 & {\cellcolor[HTML]{084B93}} \color[HTML]{F1F1F1} 0.661 & {\cellcolor[HTML]{084E98}} \color[HTML]{F1F1F1} 0.604 \\
 & TUBELEX\textsubscript{regex} & {\cellcolor[HTML]{115CA5}} \color[HTML]{F1F1F1} 0.764 & {---} & {\cellcolor[HTML]{08519C}} \color[HTML]{F1F1F1} 0.588 \\
 & TUBELEX\textsubscript{base} & {---} & {\cellcolor[HTML]{084A91}} \color[HTML]{F1F1F1} 0.663 & {---} \\
 & TUBELEX\textsubscript{lemma} & {\cellcolor[HTML]{135FA7}} \color[HTML]{F1F1F1} 0.758 & {\cellcolor[HTML]{0E59A2}} \color[HTML]{F1F1F1} 0.622 & {\cellcolor[HTML]{08458A}} \color[HTML]{F1F1F1} 0.650 \\
\hline
\multirow[c]{3}{*}{\makebox[6pt][l]{\rotatebox[origin=c]{90}{top ST}}} & Archaelogy (ID=2) & {\cellcolor[HTML]{084F99}} \color[HTML]{F1F1F1} 0.790 & {\cellcolor[HTML]{3C8CC3}} \color[HTML]{F1F1F1} 0.485 & {\cellcolor[HTML]{4594C7}} \color[HTML]{F1F1F1} 0.230 \\
 & GMU (ID=1) & {\cellcolor[HTML]{08306B}} \color[HTML]{F1F1F1} \bfseries 0.850 & {\cellcolor[HTML]{F7FBFF}} \color[HTML]{000000} 0.035 & {\cellcolor[HTML]{95C5DF}} \color[HTML]{000000} -0.073 \\
 & TMU-HIT (ID=2) & {\cellcolor[HTML]{084082}} \color[HTML]{F1F1F1} 0.820 & {\cellcolor[HTML]{08306B}} \color[HTML]{F1F1F1} \bfseries 0.733 & {\cellcolor[HTML]{08306B}} \color[HTML]{F1F1F1} \bfseries 0.762 \\
\bottomrule
\end{tabular}

    \caption{Correlation (PCC) with lexical complexity predictions, discounting misprediction of mean and variance. Best (highest) results for each language are in bold. Values for top shared task submissions (top ST) are cited from \citet{shardlow-etal-2024-bea}. See the corresponding $R^2$ results, which measure the goodness of fit, in \autoref{tab:lcp-r2}, and the discussion at the end of \autoref{sec:lcp}.} 
    \label{tab:lcp-corr}
\end{table}

\end{document}